\documentclass[runningheads]{llncs}

\usepackage[arxiv]{accv}

\usepackage{accvabbrv}

\usepackage{graphicx}
\usepackage{booktabs}
\usepackage{multirow}
\usepackage{bbm}
\usepackage[table]{xcolor}
\usepackage{subcaption}

\definecolor{rankone}{RGB}{255,235,153}    
\definecolor{ranktwo}{RGB}{224,224,224}    
\definecolor{rankthree}{RGB}{230,204,178}  
\usepackage[accsupp]{axessibility}  
\usepackage{tcolorbox}
\tcbuselibrary{breakable}

\usepackage{pifont}
\definecolor{inputheader}{RGB}{190,230,166}
\definecolor{taskheader}{RGB}{230,243,254}
\definecolor{methodheader}{RGB}{244,180,179}
\definecolor{checkgreen}{HTML}{2E7D32}
\definecolor{checkyellow}{RGB}{225,205,123}
\newcommand{\cmark}{{\color{checkgreen}\ding{51}}}
\newcommand{\pmark}{{\color{checkyellow}\ding{51}}}
\newcommand{\xmark}{{\color{gray!50}\ding{55}}}

\usepackage[pagebackref,breaklinks,colorlinks,citecolor=accvblue]{hyperref}

\usepackage{orcidlink}

\begin{document}

\title{More with Less:\\a Large Scale Remote Sensing VLM\\with a Simple Recipe} 

\titlerunning{More with Less: a Large Scale RS-VLM with a Simple Recipe}




\author{
Stefan Maria Ailuro \and
Mario Markov \and
Mohammad Mahdi \and \\
Luc Van Gool \and
Danda Pani Paudel
}

\authorrunning{S.M.~Ailuro et al.}

\institute{INSAIT, Sofia University ``St. Kliment Ohridski''\\
\email{stefan.ailuro@insait.ai}}

\maketitle

\begin{figure}
    \centering
    \includegraphics[width=1\linewidth]{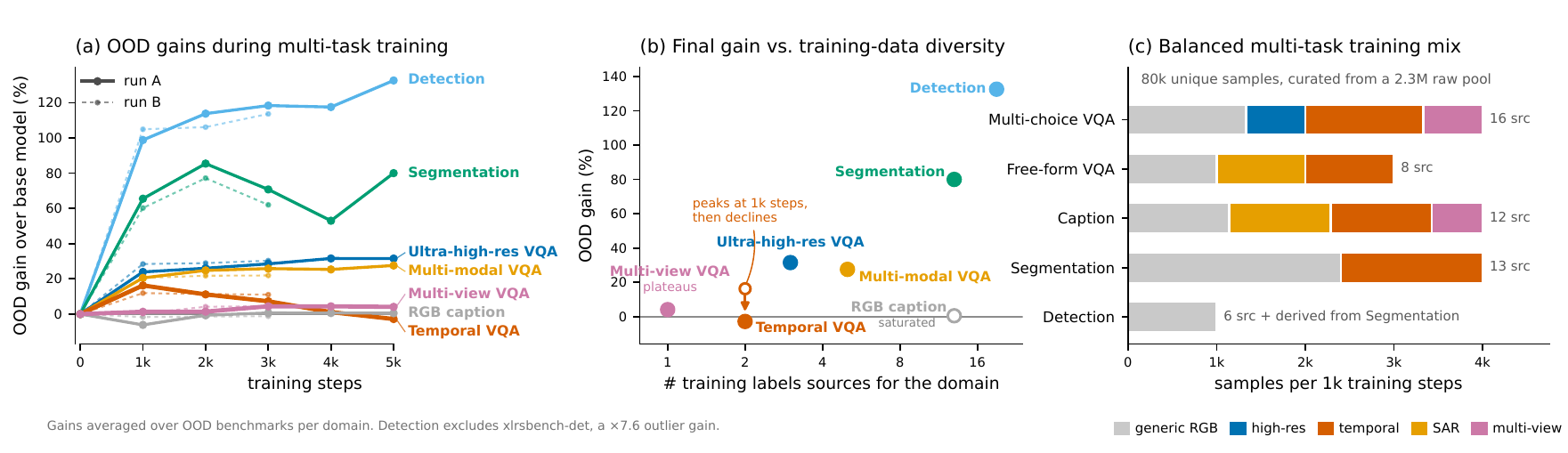}
    \caption{Diverse multi-task training drives out-of-distribution (OOD) generalization in remote sensing. (a) OOD gain over the base model during multi-task RL training, aggregated per task-domain. For robustness test the training is performed a second time with different seed and paraphrased prompts. (b) Final OOD gain vs. number of training sources per domain; for visualization, detection excludes a x7.6 gain outlier, XLRS-Bench. (c) Training mixture by task and input type; 80k samples balanced from a 2.3M pool.}
    \label{fig:teaser}
\end{figure}

\begin{abstract}
Remote sensing vision-language models are increasingly expected to support open-ended reasoning over Earth Observation data and a variety of tasks. Most recent progress in this area has been driven by remote-sensing-specific architectural designs, often introducing new encoders, alignment modules, or task-specific fusion mechanisms. In this work, we challenge the necessity of such architectural specialization. We show that a generally capable vision-language model can achieve competitive or state-of-the-art performance at challenging remote sensing benchmarks, provided that it is trained at sufficient scale across diverse data and tasks. Our model uses a single language policy that can either answer directly in text or invoke a localization tool for segmentation and grounding. To train this heterogeneous behaviour, we employ a multi-task reinforcement learning framework with adaptive task rewards covering multiple-choice VQA, free-form VQA, captioning, detection, and segmentation across a large variety of input types. Our approach achieves competitive results across a broad set of benchmarks, including high-resolution, multi-temporal, multi-modal and multi-view tasks. Further, as training data scales, our experiments show consistent improvements across most tasks both in and out of distribution, which correlate with per-task data diversity. These findings suggest that, for remote sensing VLMs, data scale is sufficient even without architectural novelty. The code and model are available at \url{https://github.com/insait-institute/MLRS}.
\end{abstract}

\section{Introduction}
\label{sec:intro}

Remote sensing is becoming central for understanding the Earth at scale. Modern satellite, aerial, and multi-sensor observation systems continuously capture information about land use, urbanization, agriculture, transportation, natural hazards, climate impacts, and environmental change. The resulting imagery is large, heterogeneous, and semantically rich: the same scene may require image-level recognition, fine-grained object localization, pixel-level segmentation, change reasoning, or open-ended question answering. As the volume and diversity of Earth observation data continue to grow, remote sensing increasingly demands models that can move beyond narrow task-specific prediction and support flexible, language-driven interaction with visual and geospatial data.

Recent progress in vision-language models has therefore motivated a new generation of remote sensing large multimodal models \cite{kuckreja2024geochat,vhm,soni2025earthdial,wang2026geollavak}. These systems aim to connect remote sensing imagery with natural language instructions, enabling unified interfaces for tasks such as captioning, visual question answering, grounding, detection, segmentation, and dialogue. A common trend in this literature is to address the peculiarities of remote sensing by designing new remote-sensing-specific architectures, alignment modules, token handling strategies, or modality-fusion mechanisms\cite{zhan2024skyeyegpt,luo2024skysensegpt,li2025lhrsbotnova,luo2025lrsvqa,shu2025earthmind,soni2025earthdial,zhang2024earthmarker,cai2026earthonevisionextendingremotesensing}. Collectively, these works demonstrate the promise of remote sensing VLMs, but they also reflect a dominant assumption: that progress in the domain requires increasingly specialized model designs.

In this work, we revisit that assumption. Instead of introducing a new remote-sensing-specific architecture, we start from a generally capable vision-language model and scale its remote sensing adaptation through data, tasks, and reinforcement learning (RL). Our central hypothesis is simple: for remote sensing VLMs, broad task coverage and large-scale training data can be sufficient even without architectural novelty. We train a single model across a large and diverse collection of remote sensing datasets and tasks, including free-form visual question answering, captioning, detection, grounding, and segmentation through promptable interaction with SAM3 \cite{carion2025sam3}. To handle the heterogeneity of these tasks, we employ a reinforcement learning strategy with adaptive per-task rewards, allowing the model to optimize across tasks with different output formats and evaluation criteria.

Our results show that this simple recipe is sufficient to achieve competitive or state-of-the-art zero-shot performance across a wide range of remote sensing benchmarks, without modifying the underlying VLM architecture. More importantly, we observe consistent gains as the amount and diversity of training data increase (see Fig. \ref{fig:teaser}), suggesting that remote sensing VLM performance remains strongly data-scalable. These findings shift the emphasis from architecture engineering toward data and task scaling: rather than designing a new model for every remote sensing capability or modality, a strong general VLM can be adapted effectively by exposing it to enough diverse remote sensing supervision and optimizing it with task-aware rewards.

Our contributions can be summarized as follows:
\begin{itemize}
    \item We present \emph{More with Less Remote Senser} (MLRS), a remote sensing VLM built without introducing any new model architecture. Despite its architectural simplicity, MLRS achieves competitive or state-of-the-art zero-shot performance across a broad range of remote sensing tasks and input types via large-scale multi-task reinforcement learning, demonstrating that a strong general-purpose VLM can be effectively adapted to compete with or surpass specialized VLMs in the remote sensing domain.

    \item We provide an empirical study of data scaling for remote sensing VLMs. Across training data sizes, model sizes, and optimization variants, we find that performance improves with increased data scale and task diversity, suggesting that data-centric scaling is a promising path toward general and capable remote sensing vision-language models.
\end{itemize}

\section{Related works}
\begin{table*}[t]
    \centering
    \caption{Capability coverage of existing RS-VLMs. UHR=Ultra-high resolution; MM=Multi-Modal (Optical, NIR, SAR); MT=Multi-Temporal; VP=Visual Prompting; VQA=Visual Question Answering; Cap=Captioning; Det=Detection; Seg=Segmentation; VG=Visual Grounding; CD=Change Detection; TS=Trained Segmentation head; RR=Reinforced Reasoning; V=Versatile, automatically identifies the task and required answer format. `\pmark'~states that segmentation is performed in textual space as a vector polygon rather than a dense mask.}
    \label{tab:rsmllmcomparison}
    \setlength{\tabcolsep}{10pt}
    \resizebox{\textwidth}{!}{
    \begin{tabular}{lcccccccccccccc}
    \toprule
    \multirow{2.4}{*}{\textbf{Model}} & \multicolumn{6}{c}{\cellcolor{taskheader}\textbf{Task Type}} & \multicolumn{3}{c}{\cellcolor{methodheader}\textbf{Method}}  & \multicolumn{5}{c}{\cellcolor{inputheader}\textbf{Input Type}}\\
    & \cellcolor{taskheader}VQA & \cellcolor{taskheader}Cap & \cellcolor{taskheader}Det & \cellcolor{taskheader}VG & \cellcolor{taskheader}Seg & \cellcolor{taskheader}CD & \cellcolor{methodheader}TS & \cellcolor{methodheader}V & \cellcolor{methodheader}RR & \cellcolor{inputheader}MT & \cellcolor{inputheader}MM & \cellcolor{inputheader}UHR & \cellcolor{inputheader}VP & \cellcolor{inputheader}MV \\ 
    \midrule
        \mbox{GeoChat~\cite{kuckreja2024geochat}} & \cmark & \cmark & \xmark & \cmark & \xmark & \xmark & \xmark & \cmark & \xmark & \xmark & \xmark & \xmark & \xmark & \xmark \\ 
        \mbox{SkySenseGPT~\cite{luo2024skysensegpt}} & \cmark & \cmark & \cmark & \cmark & \xmark & \xmark & \xmark & \cmark & \xmark & \xmark & \xmark & \xmark & \xmark & \xmark \\ 
        \mbox{LHRS-Bot-nova~\cite{li2025lhrsbotnova}} & \cmark & \cmark & \xmark & \cmark & \xmark & \xmark & \xmark & \cmark & \xmark & \xmark & \xmark & \xmark & \xmark & \xmark \\ 
        \mbox{SkyEyeGPT~\cite{zhan2024skyeyegpt}} & \cmark & \cmark & \xmark & \cmark & \xmark & \xmark & \xmark & \xmark & \xmark & \cmark & \xmark & \xmark & \xmark & \xmark \\ 
        \mbox{TEOChat~\cite{irvin2024teochat}} & \cmark & \cmark & \cmark & \cmark & \xmark & \cmark & \xmark & \cmark & \xmark & \cmark & \xmark & \xmark & \xmark & \xmark \\ 
        \mbox{DVLChat~\cite{xuan2026dynamicvl}} & \cmark & \cmark & \xmark & \xmark & \cmark & \cmark & \cmark & \xmark & \xmark & \cmark & \xmark & \xmark & \xmark & \xmark \\ 
        \mbox{VHM~\cite{vhm}} & \cmark & \cmark & \cmark & \cmark & \pmark & \xmark & \pmark & \cmark & \xmark & \xmark & \xmark & \xmark & \xmark & \xmark \\ 
        \mbox{LISAt~\cite{quenum2026lisat}} & \cmark & \cmark & \xmark & \cmark & \cmark & \xmark & \cmark & \xmark & \xmark & \xmark & \xmark & \xmark & \xmark & \xmark \\ 
        \mbox{GeoPix~\cite{geopix}} & \cmark & \cmark & \cmark & \cmark & \cmark & \xmark & \cmark & \xmark & \xmark & \xmark & \xmark & \xmark & \xmark & \xmark \\ 
        \mbox{GeoPixel~\cite{shabbir2025geopixel}} & \xmark & \cmark & \cmark & \cmark & \cmark & \xmark & \cmark & \xmark & \xmark & \xmark & \xmark & \cmark & \xmark & \xmark \\ 
        \mbox{SegEarth-R1~\cite{li2025earthreason}} & \xmark & \xmark & \xmark & \cmark & \cmark & \xmark & \cmark & \xmark & \xmark & \xmark & \xmark & \xmark & \xmark & \xmark \\ 
        \mbox{SegEarth-R2~\cite{xin2025segearthr2}} & \xmark & \xmark & \xmark & \cmark & \cmark & \xmark & \cmark & \xmark & \xmark & \xmark & \xmark & \xmark & \xmark & \xmark \\ 
        \mbox{UniGeoSeg~\cite{ni2025unigeoseg}} & \xmark & \xmark & \xmark & \cmark & \cmark & \xmark & \cmark & \xmark & \xmark & \xmark & \xmark & \xmark & \xmark & \xmark \\ 
        \mbox{Think2Seg-RS~\cite{zhang2024think2seg}} & \xmark & \xmark & \cmark & \cmark & \cmark & \xmark & \xmark & \xmark & \cmark & \xmark & \xmark & \xmark & \xmark & \xmark \\ 
        \mbox{RemoteReasoner~\cite{yao2025remotereasoner}} & \cmark & \cmark & \cmark & \cmark & \cmark & \xmark & \xmark & \cmark & \cmark & \xmark & \xmark & \xmark & \xmark & \xmark \\ 
        \mbox{RSThinker~\cite{liu2025rsthinker}} & \cmark & \cmark & \cmark & \cmark & \xmark & \xmark & \xmark & \cmark & \cmark & \xmark & \xmark & \xmark & \xmark & \xmark \\ 
        \mbox{GeoZero~\cite{wang2026geozero}} & \cmark & \cmark & \xmark & \cmark & \xmark & \xmark & \xmark & \cmark & \cmark & \xmark & \xmark & \xmark & \xmark & \xmark \\ 
        \mbox{GeoVLM-R1~\cite{fiaz2025geovlmr1}} & \cmark & \cmark & \cmark & \cmark & \xmark & \cmark & \xmark & \cmark & \cmark & \cmark & \xmark & \xmark & \xmark & \xmark \\ 
        \mbox{GeoEyes~\cite{wang2026geoeyes}} & \cmark & \xmark & \xmark & \xmark & \xmark & \xmark & \xmark & \xmark & \cmark & \cmark & \xmark & \cmark & \xmark & \xmark \\ 
        \mbox{GeoVista~\cite{zhu2026geovista}} & \cmark & \xmark & \xmark & \cmark & \xmark & \xmark & \xmark & \cmark & \cmark & \cmark & \xmark & \cmark & \xmark & \xmark \\
        \mbox{GeoLLaVA-8K~\cite{wang2026geollavak}} & \cmark & \xmark & \xmark & \xmark & \xmark & \xmark & \xmark & \xmark & \xmark & \cmark & \xmark & \cmark & \xmark & \xmark \\  
        \mbox{LRS-VQA~\cite{luo2025lrsvqa}} & \cmark & \xmark & \xmark & \cmark & \xmark & \xmark & \xmark & \cmark & \xmark & \xmark & \xmark & \cmark & \xmark & \xmark \\ 
        \mbox{UrbanLLaVA~\cite{urbanllava}} & \cmark & \xmark & \xmark & \xmark & \xmark & \xmark & \xmark & \xmark & \xmark & \xmark & \xmark & \xmark & \xmark & \cmark \\
        \mbox{EarthGPT~\cite{earthgpt}} & \cmark & \cmark & \cmark & \cmark & \xmark & \xmark & \xmark & \cmark & \xmark & \xmark & \cmark & \xmark & \xmark & \xmark \\ 
        \mbox{EarthGPT-X~\cite{earthgptx}} & \cmark & \cmark & \xmark & \xmark & \xmark & \xmark & \xmark & \cmark & \xmark & \xmark & \cmark & \xmark & \cmark & \xmark \\ 
        \mbox{EarthMaker~\cite{zhang2024earthmarker}} & \cmark & \cmark & \xmark & \xmark & \xmark & \xmark & \xmark & \cmark & \xmark & \cmark & \cmark & \xmark & \cmark & \xmark \\ 
        \mbox{EarthDial~\cite{soni2025earthdial}} & \cmark & \cmark & \cmark & \cmark & \xmark & \xmark & \xmark & \cmark & \xmark & \cmark & \cmark & \xmark & \xmark & \xmark \\ 
        \mbox{EarthMind~\cite{shu2025earthmind}} & \cmark & \cmark & \xmark & \cmark & \cmark & \cmark & \cmark & \cmark & \xmark & \cmark & \cmark & \xmark & \cmark & \xmark \\ 
        \mbox{TerraScope~\cite{shu2026terrascope}} & \cmark & \xmark & \xmark & \xmark & \cmark & \xmark & \cmark & \xmark & \xmark & \cmark & \cmark & \xmark & \xmark & \xmark \\ 
        \mbox{Falcon~\cite{yao2025falcon}} & \cmark & \cmark & \cmark & \cmark & \pmark & \cmark & \pmark & \cmark & \xmark & \cmark & \xmark & \xmark & \xmark & \xmark \\ 
        \mbox{Earth-OneVision~\cite{cai2026earthonevisionextendingremotesensing}} & \cmark & \cmark & \cmark & \cmark & \pmark & \cmark & \pmark & \cmark & \xmark & \cmark & \cmark & \xmark & \cmark & \xmark \\ 
    \midrule
        \cellcolor{rankone}\textbf{MLRS (Ours)} & \cellcolor{rankone}\cmark & \cellcolor{rankone}\cmark & \cellcolor{rankone}\cmark & \cellcolor{rankone}\cmark & \cellcolor{rankone}\cmark & \cellcolor{rankone}\cmark & \cellcolor{rankone}\cmark & \cellcolor{rankone}\cmark & \cellcolor{rankone}\cmark & \cellcolor{rankone}\cmark & \cellcolor{rankone}\cmark & \cellcolor{rankone}\cmark & \cellcolor{rankone}\cmark & \cellcolor{rankone}\cmark \\
    \bottomrule
    \end{tabular}
    }
\end{table*}

\textbf{Remote-sensing VLMs.} Adapting vision–language models to remote sensing has produced a rapidly growing family of RS-VLMs (see Tab.~\ref{tab:rsmllmcomparison}). Early instruction-tuned models such as GeoChat~\cite{kuckreja2024geochat}, SkySenseGPT~\cite{luo2024skysensegpt}, and LHRS-Bot~\cite{dilxat2025lhrsbot,li2025lhrsbotnova} established conversational VQA, captioning, and grounding on optical imagery, and subsequent work has expanded coverage along two axes: task and input data types. Along the task axis, models added detection and visual grounding~\cite{luo2024skysensegpt, irvin2024teochat}, change detection~\cite{irvin2024teochat,xuan2026dynamicvl,fiaz2025geovlmr1}, and segmentation~\cite{xuan2026dynamicvl,ni2025unigeoseg,zhang2024think2seg,quenum2026lisat,geopix,shabbir2025geopixel}. Along the input axis, dedicated architectures were proposed for multi-temporal sequences (TEOChat~\cite{irvin2024teochat}, DVLChat~\cite{xuan2026dynamicvl}, EarthDial~\cite{soni2025earthdial}), multi-modal inputs (EarthGPT~\cite{earthgpt}, EarthDial~\cite{soni2025earthdial}, EarthMind~\cite{shu2025earthmind}, TerraScope~\cite{shu2026terrascope}), ultra-high-resolution scenes (GeoLLaVA-8K~\cite{wang2026geollavak}, LRS-VQA~\cite{luo2025lrsvqa}, GeoVista~\cite{zhu2026geovista}, GeoPixel~\cite{shabbir2025geopixel}), and visual prompting (EarthMaker~\cite{zhang2024earthmarker}, EarthGPT-X~\cite{earthgptx}). Multi-view VLMs that merge information from ground views, street panoramas, and satellite imagery, are being developed separately, taking their roots in geolocalization~\cite{ye2025icrossviewgeolocalizationnatural,wang2025geovistawebaugmentedagenticvisual}, and only recently multi-view models were applied to urban analysis and broader reasoning~\cite{zhou2024urbench,urbanllava}, bringing them closer to remote sensing VLMs. A common thread in these lines of work is that each new input type is addressed with architectural specialization — temporal encoders, modality-specific branches, geometry encoders for visual prompting, or resolution-handling modules. In contrast, we show that a strong general-purpose base model trained on a sufficiently diverse multi-task, multi-domain mixture covers temporal, multi-modal, high-resolution, and multi-view inputs without architectural modification. To our knowledge, our study spans a broader combination of tasks and input types than any prior work.

\textbf{Post-training paradigms.} Most RS-VLMs are trained with supervised fine-tuning alone~\cite{kuckreja2024geochat,luo2024skysensegpt,li2025lhrsbotnova,irvin2024teochat,xuan2026dynamicvl,earthgpt,soni2025earthdial}. However, SFT-trained models tend to fit narrowly to their instruction distributions and degrade on held-out benchmarks — a pattern observed directly both in general~\cite{chu2025sftvsrl} and remote sensing~\cite{doerksen2026earthshiftbenchmarkmeasuringrobustness,ailuro2026osmdaopenstreetmapbaseddomainadaptation} domains. A recent wave of reinforcement-learning-based RS-VLMs — RSThinker~\cite{liu2025rsthinker}, GeoZero~\cite{wang2026geozero}, GeoVLM-R1~\cite{fiaz2025geovlmr1}, RemoteReasoner~\cite{yao2025remotereasoner}, Think2Seg-RS~\cite{zhang2024think2seg}, GeoEyes~\cite{wang2026geoeyes}, and GeoVista~\cite{zhu2026geovista} — reports noticeably better generalization, which our evaluation results corroborate. We therefore adopt reinforced reasoning as our post-training strategy.

\textbf{Segmentation interfaces.} RS-VLMs realize segmentation through two interfaces: emitting polygons as text (VHM~\cite{vhm}, Falcon~\cite{yao2025falcon}, Earth-OneVision~\cite{cai2026earthonevisionextendingremotesensing}) or attaching a trained segmentation head~\cite{kirillov2023sam,ravi2024sam2,cheng2021maskformer} that decodes dense masks (DVLChat~\cite{xuan2026dynamicvl}, UniGeoSeg~\cite{ni2025unigeoseg}, SegEarth~\cite{li2025earthreason,xin2025segearthr2}, LISAt~\cite{quenum2026lisat}, GeoPix~\cite{geopix}, GeoPixel~\cite{shabbir2025geopixel}, EarthMind~\cite{shu2025earthmind}, TerraScope~\cite{shu2026terrascope}). Trained dense-mask heads are better suited to remote sensing, where objects are small, numerous, and irregular~\cite{yuan2024rrsisreferringremotesensing,chen2025sam3adapter,blushteinlivnon2026samfinetuningrs}. Yet, trained segmentation heads and reinforced reasoning have been so far mutually exclusive: RL-based models either omit segmentation or rely on frozen mask decoders~\cite{zhang2024think2seg,yao2025remotereasoner}. Our model is the first remote sensing VLM to combine a trained segmentation head with reinforcement learning, via group relative tool optimization (GRTO)~\cite{markov2026bgrto}, obtaining both dense-mask quality and the generalization benefits of RL post-training.

\section{Method} 
\label{sec:method}
\begin{figure}
    \centering
    \includegraphics[width=\linewidth]{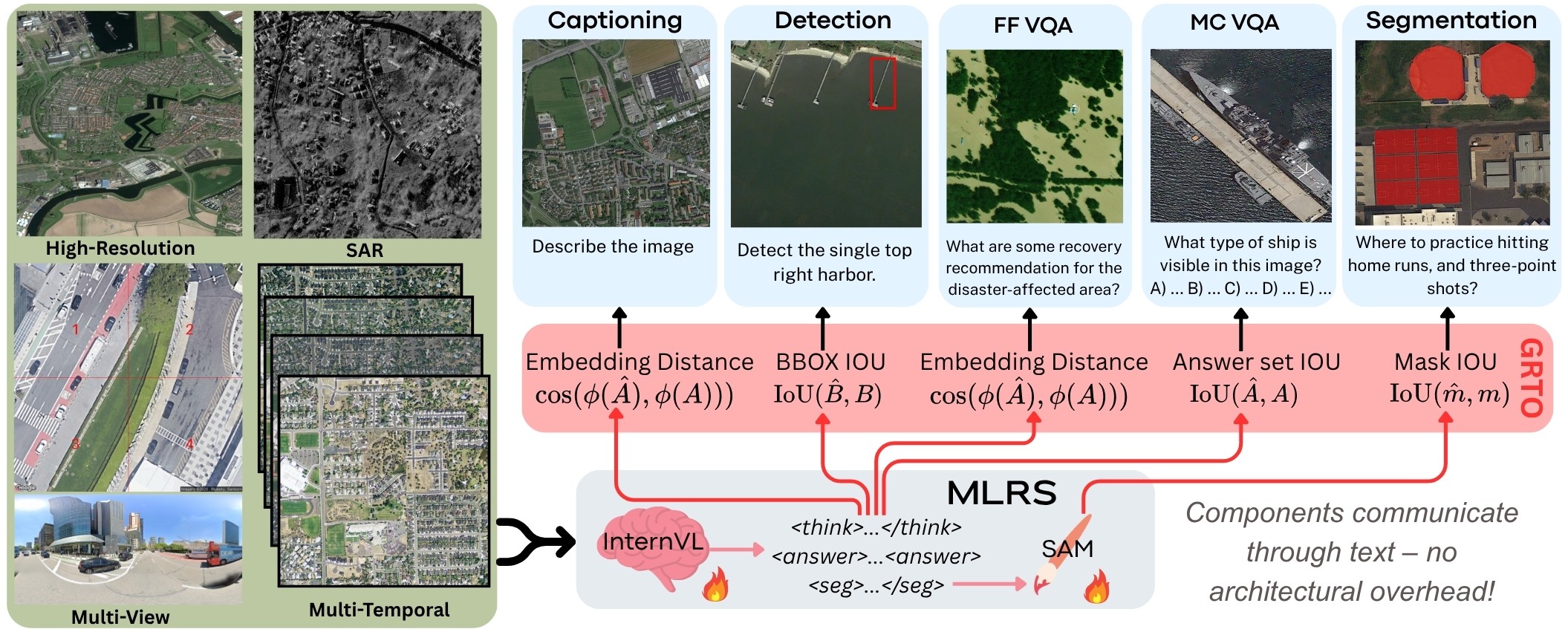}
    \caption{A single language policy (InternVL~\cite{wang2025internvl35}) handles all input types — ultra-high-resolution, multi-modal, multi-view, multi-temporal, visual prompting. The policy reasons in text, answers directly, or invokes a SAM3 segmentation tool via a textual prompt; components communicate only through text, with no task-specific encoders or fusion modules. Both components are optimized end-to-end with GRTO~\cite{markov2026bgrto}, using multi-task rewards: embedding similarity for captioning and free-form VQA, box IoU for detection, answer-set IoU for MC-VQA, and mask IoU for segmentation.}
    \label{fig:method}
\end{figure}

\textbf{Data scaling.} Our central experiment questions what drives generalization in multi-task RS training. We assemble a mixture spanning all multi-choice and free-form VQA, captioning, detection and segmentation (grounded and general)  and input types, where by construction the number and heterogeneity of sources backing each domain varies from a single dataset (multi-view) to more than a dozen (segmentation, generic VQA). We then train at scale over an increasing number of unique datapoints while holding the task composition fixed, and periodically evaluate on ID and OOD benchmarks. This design lets us measure how gains evolve with training exposure, and, crucially, whether the shape of each domain's trajectory — sustained improvement, plateau, or degradation — correlates with the diversity of its training sources rather than with the sheer volume of matching data or initial performance of the model.

\textbf{The More with Less Remote Senser (MLRS) model.}
MLRS is a tool-augmented remote sensing VLM that uses a general vision-language model as the central controller and SAM as an external localization tool. The model can operate on Optical, False-colour, SAR imagery, or mixed inputs. SAR observations are represented as grayscale images, allowing them to be processed through the same visual interface as optical images. MLRS also supports multi-image (multi-temporal, multi-view) inputs: when a task provides several images, all of them are shown to the VLM as context, while spatial outputs are produced with respect to the primary image. Finally, visual prompts are straightforwardly overlaid over the primary image.

Given the image or image group and a natural-language instruction, MLRS must decide whether the task requires a direct language response or a spatial prediction. For language-only tasks, such as VQA and captioning, it produces a textual answer. For localization tasks, it instead emits a structured segmentation tool call which describes the target regions using two complementary signals: a noun phrase and one or more bounding boxes, which are passed to SAM3~\cite{carion2025sam3}, whose output is filtered through the passed bounding boxes. In this way, the VLM is responsible for semantic grounding and coarse spatial localization, while SAM3 is responsible for producing the final pixel-level mask. This interface is similar in spirit to recent VLM--segmentation-decoder systems for referring segmentation~\cite{markov2026bgrto,yao2025remotereasoner,zhang2024think2seg}, but MLRS uses it as a general remote sensing interface for both textual answering and spatial localization.

This design keeps the VLM architecture unchanged. The model communicates with the segmentation tool only through text: it either answers directly or writes a valid localization call that can be parsed into SAM prompts. Thus, MLRS does not require new encoders, fusion blocks, alignment modules, or task-specific heads. Instead, the VLM agentically decides whether to respond in language or invoke the localization tool, enabling a single model to handle captioning, VQA, detection, and segmentation within the same interface.

\textbf{Adaptive-reward RL.} To fine-tune MLRS, we employ group relative tool optimization (GRTO)~\cite{markov2026bgrto}, where while the VLM is trained with group relative policy optimization (GRPO)~\cite{shao2024deepseekmath}, a surrogate loss is applied to the tool, SAM3, enabling joint training between policy and tool. We combine binary cross-entropy and semantic IoU for SAM3's surrogate objective, following~\cite{markov2026bgrto}.

For the GRPO policy objective, we design an adaptive task-dependent reward of the form:
\begin{equation}
R = \lambda_{\mathrm{fmt}} F + \lambda_{\mathrm{task}} S,
\label{eq:reward_general}
\end{equation}
where $F$ is a binary format-validity term, $S$ is a task-specific score, and we use $\lambda_{\mathrm{fmt}}=0.1$ and $\lambda_{\mathrm{task}}=0.9$ during training. The format term encourages the model to use the correct interface: textual tasks must produce an \texttt{answer} response, while localization tasks must produce a valid \texttt{segmentation} tool call. The score term depends on the task. For multiple-choice VQA, we parse the predicted option set and compute set IoU with the target option set, supporting both single-answer and multi-answer questions. For free-form VQA and captioning, we compare the extracted answer with the reference using Qwen3-Embedding-0.6B~\cite{zhang2025qwen3embedding}, mapping cosine similarity to $[0,1]$. For segmentation, we parse the localization call into phrase-level SAM prompts and score the resulting masks with semantic IoU against the target masks; if the ground truth contains several masks, we give zero reward if the number of predicted masks does not match, otherwise, we take the mean IoU. For detection, we parse the predicted boxes, convert them to image coordinates when necessary, greedily match them to ground-truth boxes, and score them by mean matched box IoU. Thus, the same reward template is used across all tasks, while the correctness metric changes according to the required output type. A detailed definition of each reward component is provided in the appendix. An overview of the framework is shown in Fig.~\ref{fig:method}.

\section{Experiments and results}
\subsection{Experimental Design}

\textbf{Benchmarks.} We separate evaluation into in-distribution (ID) and out-of-dis- tribution (OOD) benchmarks. ID benchmarks provide training corpora and held-out test splits to measure task learning on familiar domains: DisasterM3~\cite{wang2026disasterm3} (bi-temporal optical/SAR disaster imagery; captioning, VQA, and segmentation), DynamicVL~\cite{xuan2026dynamicvl} (long-range multi-temporal urban scenes; captioning, VQA, and segmentation over optical/SAR images), SARLANG-1M~\cite{sarlang1m} (SAR captioning and free-form VQA), EarthReason~\cite{li2025earthreason} (reasoning segmentation), LaSeRS \cite{xin2025segearthr2} (referring segmentation of varied granularity, multiplicity, linguistic variability, and reasoning requirements), and a held-out split of GeoSeg-1M -- GeoSeg-Bench \cite{ni2025unigeoseg} by Ni et al. (visual grounding and reasoning segmentation over Optical and False-colour images). OOD benchmarks share no images or annotations with training and measure generalization across all input types: XLRS-Bench~\cite{xlrsbench} probes Ultra-High-Resolution (UHR) understanding (images up to 10,000×10,000 px), reasoning, and visual prompting; RSHR-Bench~\cite{rshrbench} adds UHR MC-VQA with adversarially filtered questions and temporal reasoning; VLRS-Bench~\cite{luo2026vlrsbench} targets multi-step complex reasoning over UHR images, understanding and forecasting in multi-temporal scenes, and includes textual and visual prompting; UrBench~\cite{zhou2024urbench} evaluates multi-view reasoning over paired street-level and satellite imagery; GEOBench-VLM~\cite{geobenchvlm} spans all five tasks over multi-modal inputs (RGB, SAR, multi-spectral), and its change-detection subset additionally covers multi-temporal reasoning; and GeoSeg-Bench~\cite{jiang2026geoseg2} by Jiang et al. evaluates reasoning segmentation under varied difficulty level -- to avoid name collision with ID Benchmarks we refer to it as GeoSeg-Bench2 or GeoSeg2.

\textbf{Training mix.} To study how per-domain data diversity affects generalization, we assemble a multi-task mixture whose domains are deliberately backed by varying numbers of label sources: generic RGB tasks draw on the GeoZero corpus~\cite{wang2026geozero} (11 sources for MC-VQA, 8 for captioning, 6 for detection, 4 for free-form VQA) plus GeoLLaVA~\cite{wang2026geollavak} for high-resolution MC-VQA; segmentation combines GeoSeg-1M~\cite{ni2025unigeoseg}, LaSeRS~\cite{xin2025segearthr2}, EarthReason~\cite{li2025earthreason}, DynamicVL~\cite{xuan2026dynamicvl}, and DisasterM3~\cite{wang2026disasterm3} (13 sources); SAR tasks use SARLANG-1M~\cite{sarlang1m} and DisasterM3; multi-temporal tasks rely on only two related sources (DynamicVL, DisasterM3); and multi-view tasks on a single source. We construct multi-view instruction data from CVG-Text~\cite{ye2025icrossviewgeolocalizationnatural}, a cross-view geo-localization corpus of co-registered ground-level panoramas and satellite views: panorama captions are used directly for captioning, and we deterministically generate two MC-VQA item types — city recognition from a panorama–satellite pair, and cross-view matching, where the model must select which of four satellite views (with same-city hard negatives) depicts the panorama's location. A total raw pool of collected training data consists of 2.3M image-prompt pairs, we curate an 80k training set from it, balancing task domain and question difficulty if it is provided (see Fig.~\ref{fig:teaser}c).

\textbf{Training setup.}
We train on eight NVIDIA H200 GPUs with learning rates of $1\times10^{-6}$ for the VLM and $1\times10^{-4}$ for SAM3. We use a group size of 8, and a global batch size of 16 unique samples for a total of $16\times8=128$ rollouts per batch. We set the KL-divergence weight of the GRPO objective to $\beta=0.01$. For all trainings, we use Low-Rank Adaptation (LoRA)~\cite{hu2022lora} with rank equal to 64. The model is prompted to think, and we set the maximum output length to 1024 tokens. To balance multi-task gradients, we preserve the following composition within each batch: 25\% MC-VQA, 25\% captioning, 25\% segmentation, 18.75\% free-form VQA, and 6.25\% detection.

\textbf{Experiments.}
To study how data scale affects remote sensing performance when starting from a general-purpose VLM, we train several MLRS variants using the setup described in Section~\ref{sec:method}. Our primary run uses InternVL3.5-8B~\cite{wang2025internvl35} as the VLM backbone and is trained with GRTO for 5K steps, corresponding to 80K unique samples. To check run-to-run robustness, we train a second 8B model under the same setup for 3K steps with a different random seed and paraphrased prompt templates. To test whether the observed trends depend on model size, we also train an MLRS variant with the smaller InternVL3.5-2B backbone. We evaluate every checkpoint at 1K-step intervals, allowing us to track performance as the amount of training data increases. Finally, to disentangle the effects of data scaling between the VLM and SAM3, we train an additional 8B model for 5K steps using plain GRPO instead of GRTO, keeping SAM3 frozen. In all runs, the VLM is prompted to reason, decide whether the task requires a textual response or localization, and produce the corresponding output format. The final MLRS model refers to the last checkpoint of the primary 8B GRTO run.

\textbf{Metrics.}
We evaluate each task using metrics consistent with the corresponding benchmark whenever possible, introducing alternative metrics only when the original protocol is insufficient for our setting. For tasks with free-form answers, we use dataset-specific LLM-as-judge evaluation~\cite{zheng2023judging}. In particular, for DisasterM3~\cite{wang2026disasterm3}, DynamicVL~\cite{xuan2026dynamicvl}, and the VQA split of SARLANG-1M~\cite{sarlang1m}, we follow the dataset-specific GPT-style scoring rubrics and prompts. For GEOBench-VLM~\cite{geobenchvlm} captioning, we report BERTScore~\cite{bert-score}, following prior work. For captioning tasks in SARLANG-1M~\cite{sarlang1m} and XLRS-Bench~\cite{xlrsbench}, where prior work relies on conventional reference-overlap NLP metrics, we instead reevaluate all baselines with G-Eval~\cite{liu2023geval}, a structured LLM-as-judge framework that derives scores from next-token probability distributions rather than a single sampled rating and has been shown to align better with human judgments than reference-overlap metrics. We use Qwen2.5-32B-Instruct~\cite{qwen25} as the judge model. For multiple-choice VQA, we report accuracy. For segmentation, we report gIoU, computed as the mean per-sample intersection-over-union. For detection, we report PR@0.5 on GEOBench-VLM~\cite{geobenchvlm} and Acc@0.5 on XLRS-Bench~\cite{xlrsbench}, following the respective benchmark protocols. 

\subsection{Results}

\begin{figure}
    \centering
    \includegraphics[width=\linewidth]{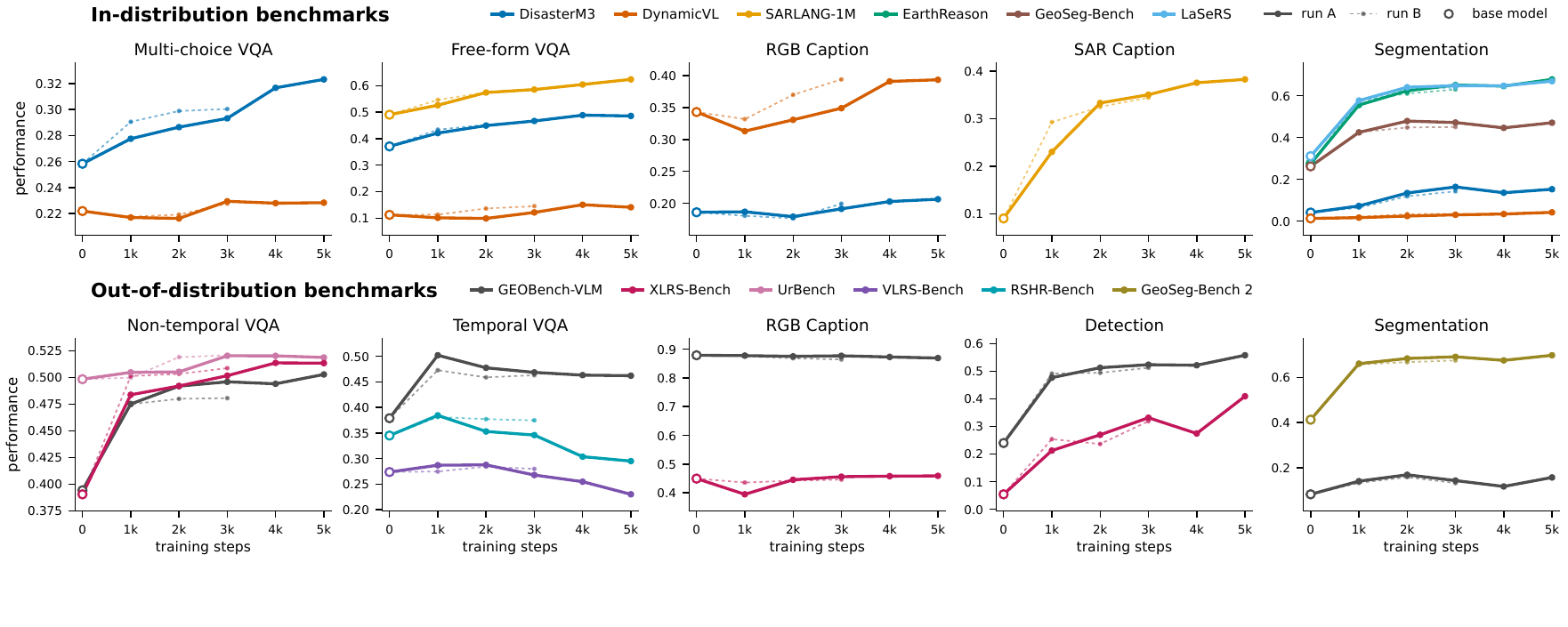}
    \caption{Performance vs training steps. Following metrics are used: aggregated accuracy for multi-choice VQA, aggregated IoU for segmentation, Precision@0.5 for GEOBench-VLM detection, BERTScore for GEOBench-VLM caption, G-Eval for other caption datasets and free-form VQA, and Accuracy@0.5 for XLRS-Bench detection.}
    \label{fig:improv}
\end{figure}

\begin{table*}[t]
    \caption{Caption, free-form VQA, and detection results. For open-source baselines, the model size closest to 8B is chosen. If multiple versions of the same baseline are reported, we take the one with the highest performance. If a baseline is fine-tuned on the corresponding benchmark, fine-tuned performance is reported and marked with an asterisk (*). Metrics in the $[0,1]$ range are rescaled to the $[0,100]$ range. In each column, the best, second-best, and third-best results are highlighted in gold, silver, and bronze, respectively. Zero-shot benchmarks are highlighted in blue.}
    \label{tab:caption-vqa-detection}
    \centering
    \resizebox{.8\linewidth}{!}{
    \begin{tabular}{lcccccccccc}
        \toprule
        & \multicolumn{5}{c}{Caption}
        & \multicolumn{3}{c}{Free-form VQA}
        & \multicolumn{2}{c}{Detection} \\
        \cmidrule(lr){2-6}
        \cmidrule(lr){7-9}
        \cmidrule(lr){10-11}
        Method
        & DM3
        & DVL
        & SL-1M
        & \cellcolor{blue!20}XLRS
        & \cellcolor{blue!20}GEO
        & DM3
        & DVL
        & SL-1M
        & \cellcolor{blue!20}GEO
        & \cellcolor{blue!20}XLRS \\
        \midrule
        Qwen
        & \cellcolor{rankone}3.90*
        & \cellcolor{ranktwo}3.03
        & 3.58
        & 39.21
        & 58.95
        & \cellcolor{ranktwo}3.11*
        & \cellcolor{ranktwo}2.34
        & \cellcolor{rankone}73.33*
        & 15.18
        & \cellcolor{rankthree}15 \\

        GPT
        & \cellcolor{ranktwo}2.57
        & \cellcolor{rankone}4.36
        & \cellcolor{rankthree}10.40
        & \cellcolor{rankone}47.49
        & \cellcolor{rankthree}64.18
        & \cellcolor{rankone}3.19
        & \cellcolor{rankone}2.75
        & -
        & 0.87
        & \cellcolor{rankone}46 \\

        GeoChat
        & -
        & -
        & 6.13
        & 3.96
        & 43.95
        & -
        & -
        & -
        & 11.51
        & 14 \\

        EarthDial
        & 1.53
        & 1.215
        & \cellcolor{ranktwo}12.04
        & 11.06
        & 53.78
        & \cellcolor{rankthree}2.42
        & \cellcolor{rankthree}1.07
        & -
        & \cellcolor{ranktwo}24.29
        & - \\

        \midrule
        Base (InternVL)
        & \cellcolor{rankthree}1.56
        & 2.49
        & 9.00
        & \cellcolor{rankthree}44.97
        & \cellcolor{rankone}87.85
        & 2.38
        & 0.355
        & \cellcolor{rankthree}25.40
        & \cellcolor{rankthree}23.96
        & 5.4 \\

        MLRS (InternVL)
        & 1.54
        & \cellcolor{rankthree}2.80
        & \cellcolor{rankone}38.24
        & \cellcolor{ranktwo}45.92
        & \cellcolor{ranktwo}86.88
        & 2.27
        & 0.810
        & \cellcolor{ranktwo}33.20
        & \cellcolor{rankone}55.74
        & \cellcolor{ranktwo}40.9 \\
        \bottomrule
    \end{tabular}
    }
\end{table*}

\begin{table*}[t]
    \caption{MC VQA and segmentation results. For open-source baselines, the model size closest to 8B is chosen. If multiple versions of the same baseline are reported, we take the one with the highest performance. If a baseline is fine-tuned on the corresponding benchmark, fine-tuned performance is reported and marked with an asterisk (*). Metrics in the $[0,1]$ range are rescaled to the $[0,100]$ range. In each column, the best, second-best, and third-best results are highlighted in gold, silver, and bronze, respectively. Zero-shot benchmarks are highlighted in blue. }
    \label{tab:mc-vqa-segmentation}
    \centering
    \begin{subtable}[b]{0.53\textwidth}
        \caption{MC VQA}
        \vspace{-3mm}
        \label{tab:mc-vqa}
        \centering
        \resizebox{\linewidth}{!}{
        \begin{tabular}{lcccccccc}
            \toprule
            Method & DM3 & DVL & \cellcolor{blue!20}RSHR & \cellcolor{blue!20}VLRS & \cellcolor{blue!20}UR & \cellcolor{blue!20}GEO & \cellcolor{blue!20}GEO (chdet) & \cellcolor{blue!20}XLRS \\
            \midrule
            Qwen & \cellcolor{rankthree}40.40* & 23.30 & 23.06 & \cellcolor{ranktwo}35.90 & - & 40.0 & 40.05 & \cellcolor{ranktwo}47.40 \\
            InternVL & \cellcolor{ranktwo}41.70* & 23.90 & 26.23 & - & \cellcolor{rankthree}48.80 & 33.3 & 26.23 & \cellcolor{rankthree}46.20 \\
            LLaVA-OneVision & 24.50 & 19.30 & - & - & - & \cellcolor{ranktwo}41.70 & 37.03 & 42.90 \\
            GeoChat & 10.70 & - & 32.12 & \cellcolor{rankthree}33.10 & - & 30.40 & - & 22.9 \\
            TEOChat & 23.00 & 17.20 & - & - & - & - & - & - \\
            EarthDial & 22.90 & \cellcolor{rankthree}30.30 & \cellcolor{ranktwo}36.99 & - & - & 37.80 & \cellcolor{rankthree}42.00 & - \\
            DVLChat & - & \cellcolor{ranktwo}33.30* & - & - & - & - & - & - \\
            GPT & \cellcolor{rankone}42.30 & \cellcolor{rankone}34.10 & \cellcolor{rankone}44.04 & \cellcolor{rankone}43.90 & \cellcolor{rankone}61.20 & \cellcolor{rankthree}41.10 & \cellcolor{ranktwo}44.04 & 38.10 \\
            \midrule
            Base (InternVL) & 25.62 & 22.18 & \cellcolor{rankthree}34.49 & 27.36 & \cellcolor{rankthree}49.83 & 39.41 & 37.89 & 39.04 \\
            MLRS (InternVL) & 32.96 & 22.83 & 29.50 & 23.00 & \cellcolor{ranktwo}51.87 & \cellcolor{rankone}50.28 & \cellcolor{rankone}46.20 & \cellcolor{rankone}51.34 \\
            \bottomrule
        \end{tabular}
        }
    \end{subtable}
    \hfill
    \begin{subtable}[b]{0.46\textwidth}
        \caption{Segmentation}
        \vspace{-3mm}
        \label{tab:segmentation}
        \centering
        \resizebox{\linewidth}{!}{
        \begin{tabular}{lccccc}
            \toprule
            Method & EarthReason & GeoSeg & LaSeRS & \cellcolor{blue!20}GeoSeg2 & \cellcolor{blue!20}GEO \\
            \midrule
            LISA & 60.88* & 4.93 & 26.6* & \cellcolor{rankthree} 39.5 & -  \\
            PixelLM & 60.01* & 4.07 & 29.3* & 0.0 & -  \\
            PSALM & \cellcolor{ranktwo} 68.30* & \cellcolor{rankthree}64.95* & - & - & -  \\
            SegEarth-R1 & \cellcolor{rankone}70.75* & \cellcolor{ranktwo}66.51* & - & - & -  \\
            UniGeoSeg & - & \cellcolor{rankone}67.75* & - & - & -  \\
            GlaMM & - & - & \cellcolor{rankthree} 43.6* & 0.0 & \cellcolor{ranktwo}14.11  \\
            SegEarth-R2 & - & - & \cellcolor{ranktwo}57.2* & - & -  \\
            \midrule
            Base (InternVL) & 27.43 & 26.2 & 31.06 & \cellcolor{ranktwo}41.29 & \cellcolor{rankthree}0.08  \\
            MLRS (InternVL) & \cellcolor{rankthree}67.78 & 47.06 & \cellcolor{rankone}66.93 & \cellcolor{rankone}69.84 & \cellcolor{rankone}15.71  \\
            \bottomrule
        \end{tabular}
        }
    \end{subtable}

\end{table*}

\begin{figure}[t]
    \centering

    \begin{subfigure}[b]{0.70\linewidth}
        \centering
        \includegraphics[width=\linewidth]{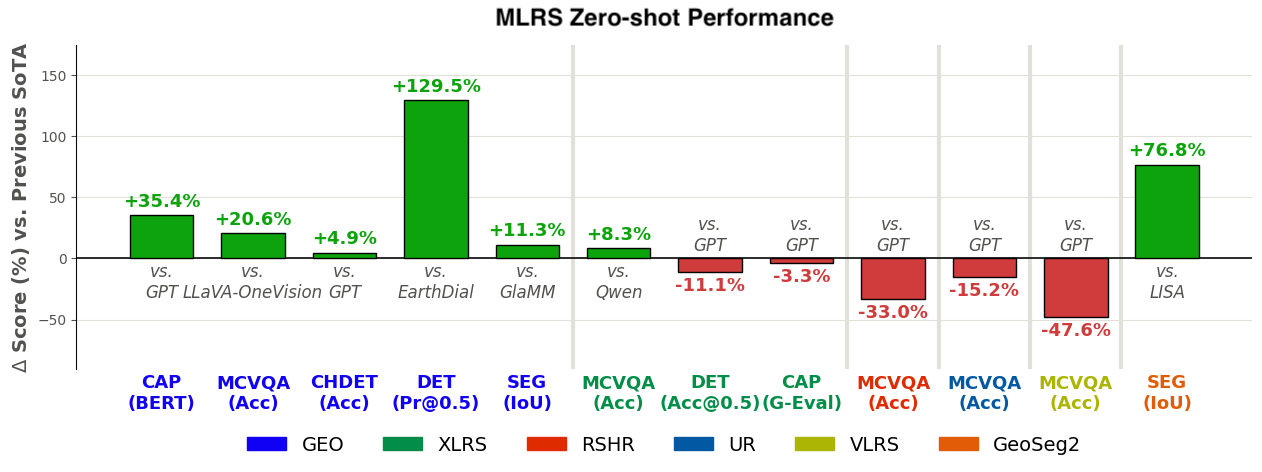}
        \caption{Zero-shot performance}
        \label{fig:zshot}
    \end{subfigure}
    \hfill
    \begin{subfigure}[b]{0.26\linewidth}
        \centering
        \includegraphics[width=\linewidth]{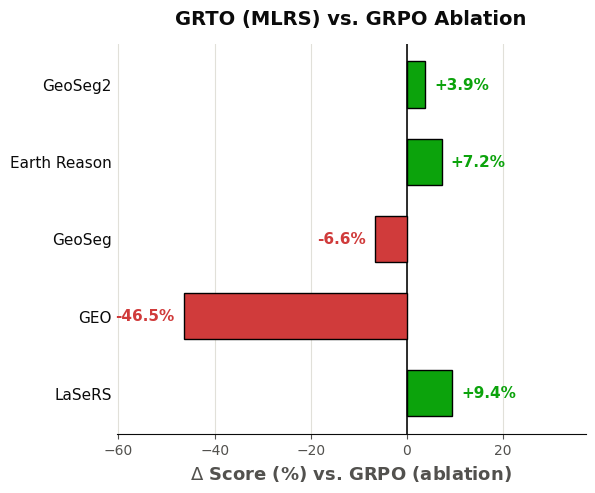}
        \caption{GRTO vs. GRPO ablation}
        \label{fig:seg-abl}
    \end{subfigure}

    \caption{Performance comparisons. (a) Zero-shot performance: percentage difference between MLRS and the best-performing existing baseline for each dataset and task. For every comparison, we report the competing model and the evaluation metric used. (b) GRTO vs. GRPO ablation: IoU percentage difference between MLRS, where SAM3 is fine-tuned (GRTO), and keeping SAM3 frozen (GRPO) for segmentation datasets.}
    \label{fig:zshot_seg_abl}
\end{figure}

\begin{figure}
    \centering
    \includegraphics[width=\linewidth]{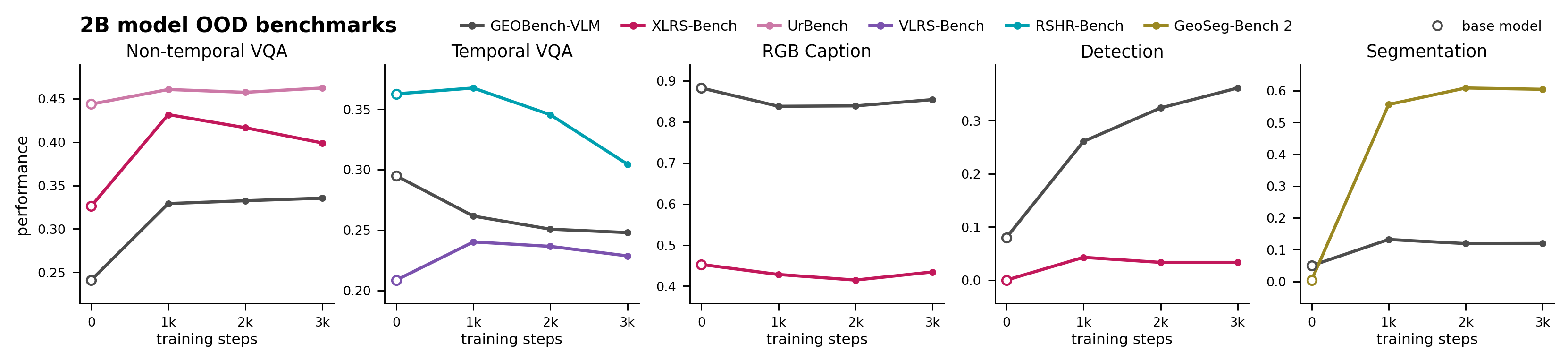}
    \caption{Model size ablation. OOD performance vs training steps of the main data scale training, reproduced for InternVL3.5-2B. The following metrics are used: aggregated accuracy for MC VQA, G-Eval for free-form VQA and captioning, aggregated IoU for segmentation, Precision@0.5 for GEOBench-VLM detection, and Accuracy@0.5 for XLRS-Bench detection.}
    \label{fig:modelsizeablation}
\end{figure}

\textbf{Effects of scaling training data.} Fig.~\ref{fig:improv} reports per-benchmark performance over the course of training for both runs. On in-distribution benchmarks, multi-task training improves majority of task families, with the largest absolute gains where the base model is weakest: SAR captioning rises from 0.09 to 0.38 G-Eval, and referring segmentation improves from 0.27 to 0.68 IoU on EarthReason and from 0.31 to 0.67 gIoU on LaSeRS. Crucially, these gains transfer out of distribution: on held-out benchmarks, detection precision on GEOBench-VLM more than doubles (0.24 → 0.56 Pr@0.5), segmentation improves from 0.41 to 0.70 IoU on GeoSeg-Bench2 and from 0.08 to 0.16 IoU on GEOBench-VLM, and VQA accuracy increases from 0.39 to 0.51 on ultra-high-resolution XLRS-Bench and from 0.39 to 0.50 on GEOBench-VLM (+28\%). XLRS-Bench detection improves by a factor of 7.6 (0.054 → 0.409 Acc@0.5). Not all domains benefit equally, however: multi-view VQA on UrBench improves only marginally (0.498 → 0.519) and plateaus after 3k steps, while temporal VQA peaks within the first 1k steps and then degrades, ending below the base model on VLRS-Bench (0.274 → 0.230) and RSHR-Bench (0.345 → 0.295). Captioning metrics on OOD benchmarks remain flat. Finally, the two 8B MLRS GRTO runs track each other closely, suggesting robustness.

In our 2B backbone ablation (see Fig.~\ref{fig:modelsizeablation} for OOD performance) the trend remains consistent: segmentation and detection continue increasing as data scales, captioning is saturated and remains relatively flat, temporal VQA, hindered by training data diversity, suffers from overfitting, while non-temporal VQA generally increases. The only exception is XLRSBench in non-temporal VQA.

\begin{figure}
    \centering
    \includegraphics[width=1\linewidth]{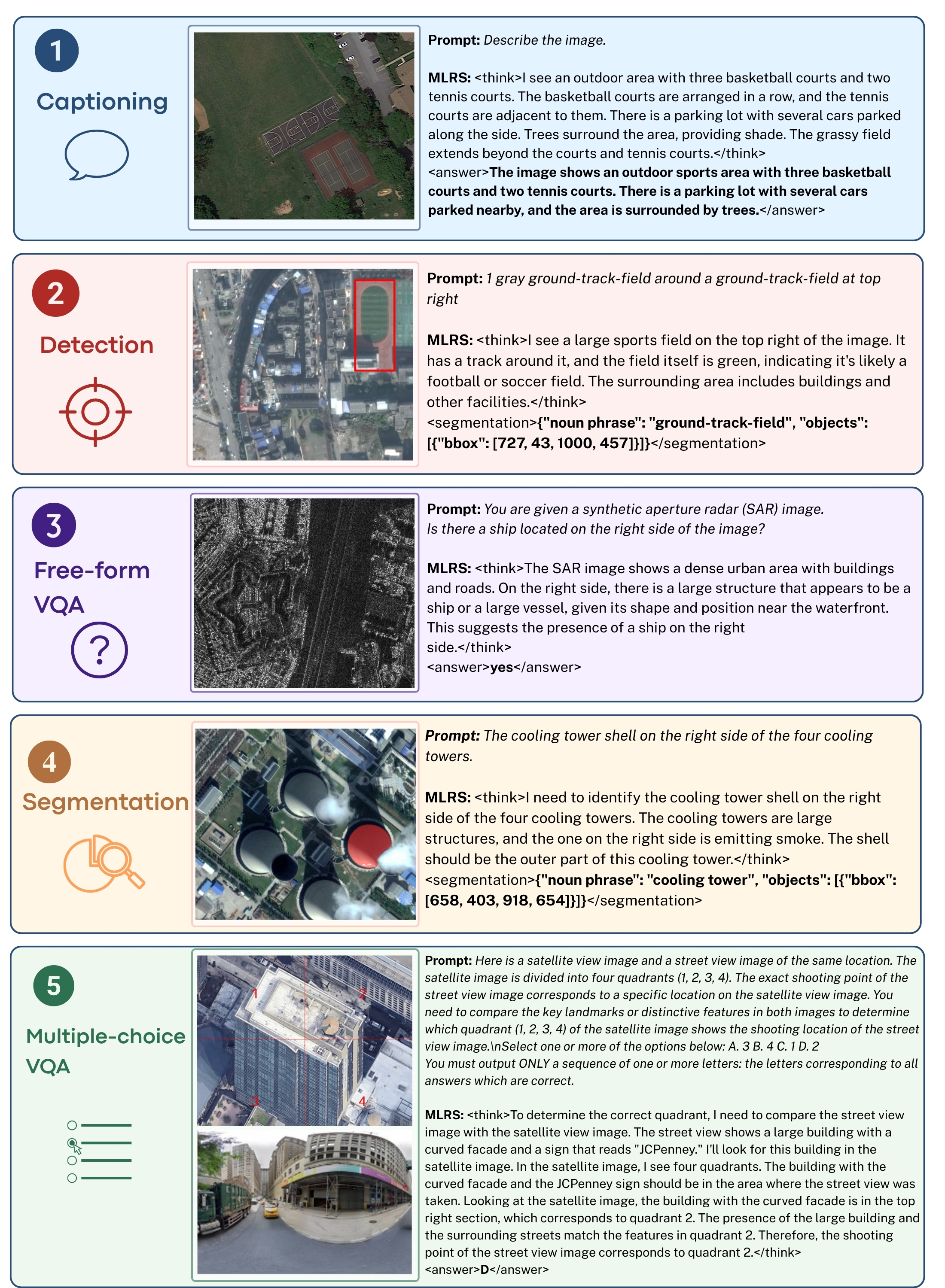}
    \caption{Example outputs of MLRS for five general tasks: captioning, detection, free-form VQA, segmentation, multiple-choice VQA. For detection, the bounding box is visualized on the input image. For segmentation, SAM3 is prompted with the VLM's output, and the resulting mask is visualized on the input image.}
    \label{fig:example_outputs}
\end{figure}

\textbf{MLRS compared to baselines.} We show the performance of MLRS compared to general-purpose and specialized VLMs in Tab.~\ref{tab:caption-vqa-detection} for captioning, free-form VQA and detection, and Tables~\ref{tab:mc-vqa} and \ref{tab:segmentation} for MC VQA and segmentation. We select the MLRS checkpoint independently of the evaluation curves: the reported model is the final checkpoint of the primary training run, evaluated uniformly across all tasks and benchmarks. In captioning, MLRS ranks second in GEOBench-VLM and XLRS-Bench -- the two zero-shot benchmarks. In the ID benchmarks, it ranks first in SARLANG-1M, and is surpassed by GPT and Qwen on DisasterM3 and DynamicVL, with a fine-tuned version of Qwen taken for the former. In free-form VQA, MLRS is overshadowed by the much larger GPT and fine-tuned versions of Qwen. In detection -- where both datasets are zero-shot -- MLRS ranks first in GEOBench-VLM and second in XLRSBench, almost surpassing GPT. In MC VQA, of the six zero-shot benchmarks, MLRS ranks first in three and second in one, failing to reach top-3 only in RSHR and VLRS. Zero-shot, it is once again overshadowed by fine-tuned baselines and GPT. Lastly, in segmentation, MLRS ranks first in both zero-shot benchmarks, first in LaSeRS, third in EarthReason, and fails to reach top-3 only in GeoSeg, where it again loses only to fine-tuned baselines. Overall, MLRS ranks first in 7 OOD benchmarks, second in 3, and fails to reach top-3 only on 2. ID, it tends to be inferior either to GPT, which is much larger, or baselines fine-tuned solely on the corresponding benchmark. Figure~\ref{fig:zshot} compares the zero-shot performance of MLRS with that of the strongest baseline models. Example outputs of MLRS for various tasks are shown in Figure~\ref{fig:example_outputs}.

\textbf{Effects of data scale on the decoder.}
Figure~\ref{fig:seg-abl} compares plain GRPO, where SAM3 is kept frozen, with GRTO, where SAM3 is fine-tuned during reinforcement learning, on the segmentation benchmarks. GRTO improves performance on more benchmarks, indicating that adapting the decoder is generally beneficial. However, the frozen-decoder GRPO variant performs substantially better on the GEOBench-VLM zero-shot benchmark, suggesting that decoder fine-tuning can reduce generalization in some settings. Since GRTO yields gains on the majority of segmentation benchmarks, we use the GRTO-trained model as the final MLRS variant.

\section{Discussion}

\textbf{Scaling and diversifying training data as a path towards state-of-the-art remote sensing.} On tasks with the most-diverse sources we observe that improvement over the base model scales well with data, and reaches competitive or state-of-the-art performance. The shape of the learning trajectory appears tied to the diversity of training data available for a domain (Fig.~\ref{fig:teaser}b): multi-view VQA, backed by a single source, plateaus at +4\%; temporal VQA, backed by only two closely related sources, peaks at +16\% after 1k steps and then declines below the base model, consistent with overfitting to a narrow temporal distribution that no longer generalizes; domains backed by three or more heterogeneous sources continue to improve throughout training. Only RGB captioning plateaus because of saturation by base model -- no headroom for improvement. Notably, the trends generally do not change when data scaling is applied to a 2B backbone, with the only exception being degrading performance in XLRSBench VQA. Possibly, the 2B model is inherently not powerful enough to process high-resolution images. Overall, this suggests that the results are not dependent on the size of the VLM.

\textbf{MLRS: all-purpose remote sensing grounded in a general-purpose architecture.} The MLRS model achieves top-1 and top-2 results on the majority of benchmarks, except multi-temporal ones, explained by a lack of training data diversity. Baselines outperform MLRS only when trained exclusively on the corresponding benchmark, but never in the OOD setting. The only exception is GPT: a vastly larger general-purpose model. Moreover, GPT is superior on benchmarks for whose annotation it was substantially used itself, i.e. DynamicVL~\cite{xuan2026dynamicvl}, RSHR-Bench~\cite{rshrbench}, XLRS-Bench caption~\cite{xlrsbench}; therefore its performance on these benchmarks could be biased. It is also notable that MLRS functions on the largest diversity of input data types and tasks across all baselines.

\textbf{Fine-tuning sparsely used components must be done with care.}
Although GRTO fine-tuning of SAM3 improves performance on a majority of segmentation benchmarks, it also leads to a substantial drop on the GEOBench-VLM zero-shot benchmark. This points to a potential overfitting risk when adapting task-specific tools, consistent with the concerns raised in B-GRTO~\cite{markov2026bgrto}. Unlike the VLM, which is exercised across all tasks and therefore receives broad and diverse supervision, SAM3 is invoked only for localization tasks and is updated using a much narrower subset of the training data. Consequently, whether such components should be fine-tuned depends not only on the amount of available data, but also on the diversity and representativeness of the subset that activates them.

\textbf{Limitations and future work.}
This study has several limitations. First, our scaling experiments use backbones from the same model family, since repeating large-scale training across many VLM families is computationally expensive. Evaluating whether the same trends hold for other general-purpose VLMs is an important direction for future work. Second, although we compare against representative set of VLMs designed for remote sensing, architecture- and data-centric approaches are yet to be directly compared under single controlled environment. Third, stronger evidence for the role of data composition would require additional controlled training runs with different data mixtures, but such experiments are costly at the scale considered here. Finally, source count is only a coarse proxy for data diversity. Given the limited number of task domains and the possibility that some domains simply have more room for improvement than others, our diversity-related trends should be interpreted as suggestive rather than conclusive. Future work should quantify intra-domain diversity more directly, for example using effective sample diversity in embedding space, and test whether such measures predict the observed peak-and-decline behavior across domains.

\section{Conclusion}
Our findings demonstrate that, given a powerful general-purpose VLM, architectural ingenuity is not necessary to achieve competitive performance in the remote sensing domain. MLRS follows a deliberately simple recipe: retain the architecture of a strong VLM, expose it to a broad mixture of remote sensing tasks and modalities, and train it to decide when to answer directly and when to invoke a localization tool. Despite this simplicity, it achieves competitive or state-of-the-art zero-shot performance across diverse benchmarks, suggesting that strong general VLMs can be effectively adapted to remote sensing through large-scale multi-task training.

More broadly, our results indicate that data scale and task diversity are central drivers of remote sensing VLM performance. As the amount and variety of training data increase, performance continues to improve, supporting a data-centric alternative to increasingly specialized architectural design. At the same time, our experiments show that tool-specific components such as SAM3 should be adapted carefully, since fine-tuning them on narrower subsets can trade in-domain gains for reduced zero-shot generalization. Overall, MLRS points toward a practical path for building general remote sensing VLMs: start from a strong general-purpose model, scale diverse remote sensing supervision, and use simple tool interfaces to extend the model to spatial tasks.

\paragraph{\textbf{Acknowledgments}}
This research was partially funded by the Ministry of Education and Science of Bulgaria (support for INSAIT, part of the Bulgarian National Roadmap for Research Infrastructure)

%
%
\bibliographystyle{splncs04}
\bibliography{main}

@String(CVPR  = {IEEE Conf. Comput. Vis. Pattern Recog.})

@String(ICCV  = {Int. Conf. Comput. Vis.})

@String(ECCV  = {Eur. Conf. Comput. Vis.})

@String(NeurIPS = {Adv. Neural Inform. Process. Syst.})

@String(AAAI  = {AAAI})

@String(CVPR  = {CVPR})

@String(ICCV  = {ICCV})

@String(ECCV  = {ECCV})

@String(NeurIPS = {NeurIPS})

@article{luo2024skysensegpt,
  title={SkySenseGPT: A Fine-Grained Instruction Tuning Dataset and Model for Remote Sensing Vision-Language Understanding},
  author={Luo, Junwei and Pang, Zhen and Zhang, Yongjun and Wang, Tingzhu and Wang, Linlin and Dang, Bo and Lao, Jiangwei and Wang, Jian and Chen, Jingdong and Tan, Yihua and Li, Yansheng},
  journal={arXiv preprint arXiv:2406.10100},
  year={2024},
  doi={10.48550/arXiv.2406.10100}
}

@article{li2025lhrsbotnova,
  title={LHRS-Bot-Nova: Improved Multimodal Large Language Model for Remote Sensing Vision-Language Interpretation},
  author={Li, Zhenshi and Muhtar, Dilxat and Gu, Feng and He, Yanglangxing and Zhang, Xueliang and Xiao, Pengfeng and He, Guangjun and Zhu, Xiaoxiang},
  journal={ISPRS Journal of Photogrammetry and Remote Sensing},
  volume={227},
  pages={539--550},
  year={2025},
  month={sep},
  doi={10.1016/j.isprsjprs.2025.06.003}
}

@article{shu2025earthmind,
  title={EarthMind: Towards Multi-Granular and Multi-Sensor Earth Observation with Large Multimodal Models},
  author={Shu, Yan and Ren, Bin and Xiong, Zhitong and Paudel, Danda Pani and Van Gool, Luc and Demir, Begum and Sebe, Nicu and Rota, Paolo},
  journal={arXiv preprint arXiv:2506.01667},
  year={2025},
  doi={10.48550/arXiv.2506.01667}
}

@inproceedings{soni2025earthdial,
  title={EarthDial: Turning Multi-sensory Earth Observations to Interactive Dialogues},
  author={Soni, Sagar and Dudhane, Akshay and Debary, Hiyam and Fiaz, Mustansar and Munir, Muhammad Akhtar and Danish, Muhammad Sohail and Fraccaro, Paolo and Watson, Campbell D. and Klein, Levente J. and Khan, Fahad Shahbaz and Khan, Salman},
  booktitle={Proceedings of the IEEE/CVF Conference on Computer Vision and Pattern Recognition (CVPR)},
  pages={14303--14313},
  month={June},
  year={2025},
  doi={10.1109/CVPR52734.2025.01334}
}

@misc{carion2025sam3,
      title={{SAM} 3: Segment Anything with Concepts},
      author={Nicolas Carion and Laura Gustafson and Yuan-Ting Hu and Shoubhik Debnath and Ronghang Hu and Didac Suris and Chaitanya Ryali and Kalyan Vasudev Alwala and Haitham Khedr and Andrew Huang and Jie Lei and Tengyu Ma and Baishan Guo and Arpit Kalla and Markus Marks and Joseph Greer and Meng Wang and Peize Sun and Roman Rädle and Triantafyllos Afouras and Effrosyni Mavroudi and Katherine Xu and Tsung-Han Wu and Yu Zhou and Liliane Momeni and Rishi Hazra and Shuangrui Ding and Sagar Vaze and Francois Porcher and Feng Li and Siyuan Li and Aishwarya Kamath and Ho Kei Cheng and Piotr Dollár and Nikhila Ravi and Kate Saenko and Pengchuan Zhang and Christoph Feichtenhofer},
      year={2025},
      eprint={2511.16719},
      archivePrefix={arXiv},
      primaryClass={cs.CV},
      url={https://arxiv.org/abs/2511.16719},
}

@article{zhang2025qwen3embedding,
  title         = {Qwen3 Embedding: Advancing Text Embedding and Reranking Through Foundation Models},
  author        = {Zhang, Yanzhao and Li, Mingxin and Long, Dingkun and Zhang, Xin and Lin, Huan and Yang, Baosong and Xie, Pengjun and Yang, An and Liu, Dayiheng and Lin, Junyang and Huang, Fei and Zhou, Jingren},
  journal       = {arXiv preprint arXiv:2506.05176},
  year          = {2025},
  doi           = {10.48550/arXiv.2506.05176},
  archivePrefix = {arXiv},
  eprint        = {2506.05176},
  primaryClass  = {cs.CL}
}

@misc{shao2024deepseekmath,
  author = {Zhihong Shao and Peiyi Wang and Qihao Zhu and Runxin Xu and Junxiao Song and Mingchuan Zhang and Y.K. Li and Y. Wu and Daya Guo},
  title = {{DeepSeekMath}: Pushing the Limits of Mathematical Reasoning in Open Language Models},
  journal = {CoRR},
  volume = {abs/2402.03300},
  year = {2024},
  url = {https://arxiv.org/abs/2402.03300},
}

@article{markov2026bgrto,
  title         = {B-GRTO: Bootstrapped Group Relative Tool Optimization for Referring Segmentation},
  author        = {Markov, Mario and Ailuro, Stefan Maria and Mahdi, Mohammad and Van Gool, Luc and Paudel, Danda Pani},
  journal       = {arXiv preprint arXiv:2605.23500},
  year          = {2026},
  doi           = {10.48550/arXiv.2605.23500},
  archivePrefix = {arXiv},
  eprint        = {2605.23500},
  primaryClass  = {cs.CV}
}

@inproceedings{zheng2023judging,
  title     = {Judging {LLM}-as-a-Judge with {MT}-Bench and Chatbot Arena},
  author    = {Zheng, Lianmin and Chiang, Wei-Lin and Sheng, Ying and Zhuang, Siyuan and Wu, Zhanghao and Zhuang, Yonghao and Lin, Zi and Li, Zhuohan and Li, Dacheng and Xing, Eric P. and Zhang, Hao and Gonzalez, Joseph E. and Stoica, Ion},
  booktitle = {Advances in Neural Information Processing Systems},
  volume    = {36},
  pages     = {46595--46623},
  year      = {2023}
}

@article{qwen25,
    title   = {Qwen2.5 Technical Report}, 
    author  = {An Yang and Baosong Yang and Beichen Zhang and Binyuan Hui and Bo Zheng and Bowen Yu and Chengyuan Li and Dayiheng Liu and Fei Huang and Haoran Wei and Huan Lin and Jian Yang and Jianhong Tu and Jianwei Zhang and Jianxin Yang and Jiaxi Yang and Jingren Zhou and Junyang Lin and Kai Dang and Keming Lu and Keqin Bao and Kexin Yang and Le Yu and Mei Li and Mingfeng Xue and Pei Zhang and Qin Zhu and Rui Men and Runji Lin and Tianhao Li and Tingyu Xia and Xingzhang Ren and Xuancheng Ren and Yang Fan and Yang Su and Yichang Zhang and Yu Wan and Yuqiong Liu and Zeyu Cui and Zhenru Zhang and Zihan Qiu},
    journal = {arXiv preprint arXiv:2412.15115},
    year    = {2024}
}

@article{wang2025internvl35,
  title   = {InternVL3.5: Advancing Open-Source Multimodal Models in Versatility, Reasoning, and Efficiency},
  author  = {Wang, Weiyun and Gao, Zhangwei and Gu, Lixin and Pu, Hengjun and Cui, Long and Wei, Xingguang and Liu, Zhaoyang and Jing, Linglin and Ye, Shenglong and Shao, Jie and others},
  journal = {arXiv preprint arXiv:2508.18265},
  year    = {2025},
  doi     = {10.48550/arXiv.2508.18265},
  url     = {https://arxiv.org/abs/2508.18265}
}

@inproceedings{
hu2022lora,
title={Lo{RA}: Low-Rank Adaptation of Large Language Models},
author={Edward J Hu and yelong shen and Phillip Wallis and Zeyuan Allen-Zhu and Yuanzhi Li and Shean Wang and Lu Wang and Weizhu Chen},
booktitle={International Conference on Learning Representations},
year={2022},
url={https://openreview.net/forum?id=nZeVKeeFYf9}
}

@inproceedings{liu2023geval,
    title = "G-{E}val: {NLG} Evaluation using {GPT}-4 with Better Human Alignment",
    author = "Liu, Yang  and
      Iter, Dan  and
      Xu, Yichong  and
      Wang, Shuohang  and
      Xu, Ruochen  and
      Zhu, Chenguang",
    editor = "Bouamor, Houda  and
      Pino, Juan  and
      Bali, Kalika",
    booktitle = "Proceedings of the 2023 Conference on Empirical Methods in Natural Language Processing",
    month = dec,
    year = "2023",
    address = "Singapore",
    publisher = "Association for Computational Linguistics",
    url = "https://aclanthology.org",
    doi = "10.18653/v1/2023.emnlp-main.153",
    pages = "2511--2522"
}

@article{vhm, 
    title={VHM: Versatile and Honest Vision Language Model for Remote Sensing Image Analysis}, 
    volume={39}, 
    url={https://ojs.aaai.org/index.php/AAAI/article/view/32683}, 
    DOI={10.1609/aaai.v39i6.32683},
    number={6}, 
    journal={Proceedings of the AAAI Conference on Artificial Intelligence}, 
    author={Pang, Chao and Weng, Xingxing and Wu, Jiang and Li, Jiayu and Liu, Yi and Sun, Jiaxing and Li, Weijia and Wang, Shuai and Feng, Litong and Xia, Gui-Song and He, Conghui}, 
    year={2025}, 
    month={Apr.}, 
    pages={6381-6388} 
}

@InProceedings{xlrsbench,
    author    = {Wang, Fengxiang and Wang, Hongzhen and Guo, Zonghao and Wang, Di and Wang, Yulin and Chen, Mingshuo and Ma, Qiang and Lan, Long and Yang, Wenjing and Zhang, Jing and Liu, Zhiyuan and Sun, Maosong},
    title     = {XLRS-Bench: Could Your Multimodal LLMs Understand Extremely Large Ultra-High-Resolution Remote Sensing Imagery?},
    booktitle = {Proceedings of the IEEE/CVF Conference on Computer Vision and Pattern Recognition (CVPR)},
    month     = {June},
    year      = {2025},
    pages     = {14325-14336}
}

@article{zhan2024skyeyegpt,
title = {SkyEyeGPT: Unifying remote sensing vision-language tasks via instruction tuning with large language model},
journal = {ISPRS Journal of Photogrammetry and Remote Sensing},
volume = {221},
pages = {64-77},
year = {2025},
issn = {0924-2716},
doi = {https://doi.org/10.1016/j.isprsjprs.2025.01.020},
url = {https://www.sciencedirect.com/science/article/pii/S0924271625000206},
author = {Yang Zhan and Zhitong Xiong and Yuan Yuan},
}

@inproceedings{kuckreja2024geochat,
    author    = {Kuckreja, Kartik and Danish, Muhammad Sohail and Naseer, Muzammal and Das, Abhijit and Khan, Salman and Khan, Fahad Shahbaz},
    title     = {GeoChat: Grounded Large Vision-Language Model for Remote Sensing},
    booktitle = {Proceedings of the IEEE/CVF Conference on Computer Vision and Pattern Recognition (CVPR)},
    month     = {June},
    year      = {2024},
    pages     = {27831-27840}
}

@InProceedings{luo2025lrsvqa,
    title={When Large Vision-Language Model Meets Large Remote Sensing Imagery: Coarse-to-Fine Text-Guided Token Pruning},
    author={Luo, Junwei and Zhang, Yingying and Yang, Xue and Wu, Kang and Zhu, Qi and Liang, Lei and Chen, Jingdong and Li, Yansheng},
    booktitle={Proceedings of the IEEE/CVF International Conference on Computer Vision (ICCV)},
    month={October},
    year={2025},
    pages={9206-9217}
}

@ARTICLE{geopix,
  author={Ou, Ruizhe and Hu, Yuan and Zhang, Fan and Chen, Jiaxin and Liu, Yu},
  journal={IEEE Geoscience and Remote Sensing Magazine}, 
  title={GeoPix: A multimodal large language model for pixel-level image understanding in remote sensing}, 
  year={2025},
  volume={13},
  number={3},
  pages={324-337},
  doi={10.1109/MGRS.2025.3560293}
}

@ARTICLE{earthgpt,
  author={Zhang, Wei and Cai, Miaoxin and Zhang, Tong and Zhuang, Yin and Mao, Xuerui},
  journal={IEEE Transactions on Geoscience and Remote Sensing}, 
  title={EarthGPT: A Universal Multimodal Large Language Model for Multisensor Image Comprehension in Remote Sensing Domain}, 
  year={2024},
  volume={62},
  number={},
  pages={1-20},
  doi={10.1109/TGRS.2024.3409624}
}

@InProceedings{dilxat2025lhrsbot,
author="Muhtar, Dilxat
and Li, Zhenshi
and Gu, Feng
and Zhang, Xueliang
and Xiao, Pengfeng",
editor="Leonardis, Ale{\v{s}}
and Ricci, Elisa
and Roth, Stefan
and Russakovsky, Olga
and Sattler, Torsten
and Varol, G{\"u}l",
title="LHRS-Bot: Empowering Remote Sensing with VGI-Enhanced Large Multimodal Language Model",
booktitle="Computer Vision -- ECCV 2024",
year="2025",
publisher="Springer Nature Switzerland",
address="Cham",
pages="440--457",
isbn="978-3-031-72904-1"
}

@misc{wang2026geozero,
      title={GeoZero: Incentivizing Reasoning from Scratch on Geospatial Scenes}, 
      author={Di Wang and Shunyu Liu and Wentao Jiang and Fengxiang Wang and Yi Liu and Xiaolei Qin and Zhiming Luo and Chaoyang Zhou and Haonan Guo and Jing Zhang and Bo Du and Dacheng Tao and Liangpei Zhang},
      year={2026},
      eprint={2511.22645},
      archivePrefix={arXiv},
      primaryClass={cs.CV},
      url={https://arxiv.org/abs/2511.22645}, 
}

@misc{liu2025rsthinker,
      title={Towards Faithful Reasoning in Remote Sensing: A Perceptually-Grounded GeoSpatial Chain-of-Thought for Vision-Language Models}, 
      author={Jiaqi Liu and Lang Sun and Ronghao Fu and Bo Yang},
      year={2026},
      eprint={2509.22221},
      archivePrefix={arXiv},
      primaryClass={cs.CV},
      url={https://arxiv.org/abs/2509.22221}, 
}

@article{kirillov2023sam,
  title={Segment Anything},
  author={Kirillov, Alexander and Mintun, Eric and Ravi, Nikhila and Mao, Hanzi and Rolland, Chloe and Gustafson, Laura and Xiao, Tete and Whitehead, Spencer and Berg, Alexander C. and Lo, Wan-Yen and Doll{\'a}r, Piotr and Girshick, Ross},
  journal={arXiv:2304.02643},
  year={2023}
}

@article{zhang2024think2seg,
title = {Bridging semantics and geometry: A decoupled LVLM–SAM framework for reasoning segmentation in optical remote sensing},
journal = {ISPRS Journal of Photogrammetry and Remote Sensing},
volume = {237},
pages = {217-235},
year = {2026},
issn = {0924-2716},
doi = {https://doi.org/10.1016/j.isprsjprs.2026.04.036},
url = {https://www.sciencedirect.com/science/article/pii/S0924271626002091},
author = {Xu Zhang and Junyao Ge and Yang Zheng and Kaitai Guo and Jimin Liang},
}

@article{ravi2024sam2,
  title={{SAM} 2: Segment Anything in Images and Videos},
  author={Ravi, Nikhila and Gabeur, Valentin and Hu, Yuan-Ting and Hu, Ronghang and Ryali, Chaitanya and Ma, Tengyu and Khedr, Haitham and R{\"a}dle, Roman and Rolland, Chloe and Gustafson, Laura and Mintun, Eric and Pan, Junting and Alwala, Kalyan Vasudev and Carion, Nicolas and Wu, Chao-Yuan and Girshick, Ross and Doll{\'a}r, Piotr and Feichtenhofer, Christoph},
  journal={arXiv preprint arXiv:2408.00714},
  url={https://arxiv.org/abs/2408.00714},
  year={2024}
}

@misc{chen2025sam3adapter,
      title={{SAM3-Adapter}: Efficient Adaptation of Segment Anything 3 for Camouflage Object Segmentation, Shadow Detection, and Medical Image Segmentation}, 
      author={Tianrun Chen and Runlong Cao and Xinda Yu and Lanyun Zhu and Chaotao Ding and Deyi Ji and Cheng Chen and Qi Zhu and Chunyan Xu and Papa Mao and Ying Zang},
      year={2025},
      eprint={2511.19425},
      archivePrefix={arXiv},
      primaryClass={cs.CV},
      url={https://arxiv.org/abs/2511.19425}, 
}

@misc{li2025earthreason,
  title         = {{SegEarth-R1}: Geospatial Pixel Reasoning via Large Language Model},
  author        = {Li, Kaiyu and Xin, Zepeng and Pang, Li and Pang, Chao and Deng, Yupeng and Yao, Jing and Xia, Guisong and Meng, Deyu and Wang, Zhi and Cao, Xiangyong},
  year          = {2025},
  eprint        = {2504.09644},
  doi           = {10.48550/arXiv.2504.09644},
  url           = {https://arxiv.org/abs/2504.09644}
}

@misc{xin2025segearthr2,
      title={SegEarth-R2: Towards Comprehensive Language-guided Segmentation for Remote Sensing Images}, 
      author={Zepeng Xin and Kaiyu Li and Luodi Chen and Wanchen Li and Yuchen Xiao and Hui Qiao and Weizhan Zhang and Deyu Meng and Xiangyong Cao},
      year={2025},
      eprint={2512.20013},
      archivePrefix={arXiv},
      primaryClass={cs.CV},
      url={https://arxiv.org/abs/2512.20013}, 
}

@misc{ni2025unigeoseg,
      title={UniGeoSeg: Towards Unified Open-World Segmentation for Geospatial Scenes}, 
      author={Shuo Ni and Di Wang and He Chen and Haonan Guo and Ning Zhang and Jing Zhang},
      year={2025},
      eprint={2511.23332},
      archivePrefix={arXiv},
      primaryClass={cs.CV},
      url={https://arxiv.org/abs/2511.23332}, 
}

@inproceedings{yao2025remotereasoner,
  title={Remotereasoner: Towards unifying geospatial reasoning workflow},
  author={Yao, Liang and Liu, Fan and Lu, Hongbo and Zhang, Chuanyi and Min, Rui and Xu, Shengxiang and Di, Shimin and Peng, Pai},
  booktitle={Proceedings of the AAAI Conference on Artificial Intelligence},
  volume={40},
  number={14},
  pages={11883--11891},
  year={2026}
}

@article{shabbir2025geopixel,
  title={Geopixel: Pixel grounding large multimodal model in remote sensing},
  author={Shabbir, Akashah and Zumri, Mohammed and Bennamoun, Mohammed and Khan, Fahad S and Khan, Salman},
  journal={arXiv preprint arXiv:2501.13925},
  year={2025}
}

@inproceedings{
xuan2026dynamicvl,
title={Dynamic{VL}: Benchmarking Multimodal Large Language Models for Dynamic City Understanding},
author={Weihao Xuan and Junjue Wang and Heli Qi and Zihang Chen and Zhuo Zheng and Yanfei Zhong and Junshi Xia and Naoto Yokoya},
booktitle={The Thirty-ninth Annual Conference on Neural Information Processing Systems Datasets and Benchmarks Track},
year={2026},
url={https://openreview.net/forum?id=zubCrOvUZ4}
}

@inproceedings{irvin2024teochat,
  title={TEOChat: A Large Vision-Language Assistant for Temporal Earth Observation Data},
  author={Irvin, Jeremy Andrew and Liu, Emily Ruoyu and Chen, Joyce Chuyi and Dormoy, Ines and Kim, Jinyoung and Khanna, Samar and Zheng, Zhuo and Ermon, Stefano},
  booktitle={International Conference on Learning Representations},
  year={2025}
}

@misc{fiaz2025geovlmr1,
      title={GeoVLM-R1: Reinforcement Fine-Tuning for Improved Remote Sensing Reasoning}, 
      author={Mustansar Fiaz and Hiyam Debary and Paolo Fraccaro and Danda Paudel and Luc Van Gool and Fahad Khan and Salman Khan},
      year={2025},
      eprint={2509.25026},
      archivePrefix={arXiv},
      primaryClass={cs.CV},
      url={https://arxiv.org/abs/2509.25026}, 
}

@inproceedings{
quenum2026lisat,
title={{LISA}t: Language-Instructed Segmentation Assistant for Satellite Imagery},
author={Jerome Quenum and Wen-Han Hsieh and Tsung-Han Wu and Ritwik Gupta and Trevor Darrell and David M. Chan},
booktitle={The Thirty-ninth Annual Conference on Neural Information Processing Systems Datasets and Benchmarks Track},
year={2026},
url={https://openreview.net/forum?id=X7CxMmmgkb}
}

@inproceedings{
wang2026geollavak,
title={Geo{LL}a{VA}-8K: Scaling Remote-Sensing Multimodal Large Language Models to 8K Resolution},
author={Fengxiang Wang and Mingshuo Chen and Yueying Li and Di Wang and Haotian Wang and Zonghao Guo and Zefan Wang and Shan Boqi and Long Lan and Yulin Wang and Hongzhen Wang and Wenjing Yang and Bo Du and Jing Zhang},
booktitle={The Thirty-ninth Annual Conference on Neural Information Processing Systems},
year={2026},
url={https://openreview.net/forum?id=LTgUInLTbP}
}

@inproceedings{shu2026terrascope,
  title={TerraScope: Pixel-Grounded Visual Reasoning for Earth Observation},
  author={Shu, Yan and Ren, Bin and Xiong, Zhitong and Zhu, Xiao Xiang and Demir, Beg{\"u}m and Sebe, Nicu and Rota, Paolo},
  booktitle={IEEE conference on Computer Vision and Pattern Recognition},
  year={2026}
}

@article{wang2026geoeyes,
  title={GeoEyes: On-Demand Visual Focusing for Evidence-Grounded Understanding of Ultra-High-Resolution Remote Sensing Imagery},
  author={Wang, Fengxiang and Chen, Mingshuo and Li, Yueying and Yang, Yajie and Zhang, Yifan and Lan, Long and Yang, Xue and Sun, Hongda and Wang, Yulin and Wang, Di and others},
  journal={arXiv preprint arXiv:2602.14201},
  year={2026}
}

@misc{zhu2026geovista,
      title={GeoVista: Visually Grounded Active Perception for Ultra-High-Resolution Remote Sensing Understanding}, 
      author={Zhu, Jiashun and Fu, Ronghao and Hu, Jiasen and Xing. Nachuan et al.},
      year={2026},
      eprint={2605.14475},
      archivePrefix={arXiv},
      primaryClass={cs.CV},
      url={https://arxiv.org/abs/2605.14475}, 
}

@article{zhang2024earthmarker,
  title={EarthMarker: A Visual Prompting Multi-modal Large Language Model for Remote Sensing},
  author={Zhang, Wei and Cai, Miaoxin and Zhang, Tong and Zhuang, Yin and Li, Jun and Mao, Xuerui},
  journal={IEEE Transactions on Geoscience and Remote Sensing},
  year={2024},
  publisher={IEEE}
}

@ARTICLE{earthgptx,
  author={Zhang, Wei and Cai, Miaoxin and Ning, Yaqian and Zhang, Tong and Zhuang, Yin and Lu, Shijian and Chen, He and Li, Jun and Mao, Xuerui},
  journal={IEEE Transactions on Geoscience and Remote Sensing}, 
  title={EarthGPT-X: A Spatial MLLM for Multilevel Multisource Remote Sensing Imagery Understanding With Visual Prompting}, 
  year={2025},
  volume={63},
  number={},
  pages={1-21},
  doi={10.1109/TGRS.2025.3626941}
}

@misc{yao2025falcon,
  title={Falcon: A Remote Sensing Vision-Language Foundation Model},
  author={kelu, Yao and Nuo, Xu and Rong, Yang and Yingying, Xu and Zhuoyan, Gao and Titinunt, Kitrungrotsakul and yi, Ren and Pu, Zhang and Jin, Wang and Ning, Wei and Chao, Li},
  journal={arXiv preprint arXiv:2503.11070},
  year={2025}
}

@misc{cai2026earthonevisionextendingremotesensing,
      title={Earth-OneVision: Extending Remote Sensing Multimodal Large Language Models to More Sensor Modalities and Tasks}, 
      author={Miaoxin Cai and Guanqun Wang and Wei Zhang and Guangyao Zhou and Yin Zhuang and Tong Zhang and Hao Wang and He Chen and Jun Li},
      year={2026},
      eprint={2606.10819},
      archivePrefix={arXiv},
      primaryClass={cs.CV},
      url={https://arxiv.org/abs/2606.10819}, 
}

@misc{ye2025icrossviewgeolocalizationnatural,
      title={Where am I? Cross-View Geo-localization with Natural Language Descriptions}, 
      author={Junyan Ye and Honglin Lin and Leyan Ou and Dairong Chen and Zihao Wang and Qi Zhu and Conghui He and Weijia Li},
      year={2025},
      eprint={2412.17007},
      archivePrefix={arXiv},
      primaryClass={cs.CV},
      url={https://arxiv.org/abs/2412.17007}, 
}

@misc{wang2025geovistawebaugmentedagenticvisual,
      title={GeoVista: Web-Augmented Agentic Visual Reasoning for Geolocalization}, 
      author={Yikun Wang and Zuyan Liu and Ziyi Wang and Han Hu and Pengfei Liu and Yongming Rao},
      year={2025},
      eprint={2511.15705},
      archivePrefix={arXiv},
      primaryClass={cs.CV},
      url={https://arxiv.org/abs/2511.15705}, 
}

@InProceedings{urbanllava,
    author    = {Feng, Jie and Wang, Shengyuan and Liu, Tianhui and Xi, Yanxin and Li, Yong},
    title     = {UrbanLLaVA: A Multi-modal Large Language Model for Urban Intelligence},
    booktitle = {Proceedings of the IEEE/CVF International Conference on Computer Vision (ICCV)},
    month     = {October},
    year      = {2025},
    pages     = {6209-6219}
}

@inproceedings{cheng2021maskformer,
  title={Per-Pixel Classification is Not All You Need for Semantic Segmentation},
  author={Bowen Cheng and Alexander G. Schwing and Alexander Kirillov},
  journal={NeurIPS},
  booktitle={NeurIPS},
  year={2021}
}

@inproceedings{
chu2025sftvsrl,
title={{SFT} Memorizes, {RL} Generalizes: A Comparative Study of Foundation Model Post-training},
author={Tianzhe Chu and Yuexiang Zhai and Jihan Yang and Shengbang Tong and Saining Xie and Dale Schuurmans and Quoc V Le and Sergey Levine and Yi Ma},
booktitle={Forty-second International Conference on Machine Learning},
year={2025},
url={https://openreview.net/forum?id=dYur3yabMj}
}

@article{zhou2024urbench,
  title={Urbench: A comprehensive benchmark for evaluating large multimodal models in multi-view urban scenarios},
  author={Zhou, Baichuan and Yang, Haote and Chen, Dairong and Ye, Junyan and Bai, Tianyi and Yu, Jinhua and Zhang, Songyang and Lin, Dahua and He, Conghui and Li, Weijia},
  journal={AAAI},
  year={2025}
}

@misc{ailuro2026osmdaopenstreetmapbaseddomainadaptation,
      title={OSMDA: OpenStreetMap-based Domain Adaptation for Remote Sensing VLMs}, 
      author={Stefan Maria Ailuro and Mario Markov and Mohammad Mahdi and Delyan Boychev and Luc Van Gool and Danda Pani Paudel},
      year={2026},
      eprint={2603.11804},
      archivePrefix={arXiv},
      primaryClass={cs.CV},
      url={https://arxiv.org/abs/2603.11804}, 
}

@misc{doerksen2026earthshiftbenchmarkmeasuringrobustness,
  title     = {EarthShift: a benchmark for measuring robustness to real-world distribution shifts in Earth observation},
  author    = {Kelsey Doerksen and Hannah Kerner},
  eprint   = {2605.29330},
  year      = {2026},
  archivePrefix={arXiv},
  primaryClass={cs.CV},
  url={https://arxiv.org/abs/2605.29330}
}

@misc{blushteinlivnon2026samfinetuningrs,
      title={On the Effectiveness of Textual Prompting with Lightweight Fine-Tuning for SAM3 Remote Sensing Segmentation}, 
      author={Roni Blushtein-Livnon and Osher Rafaeli and David Ioffe and Amir Boger and Karen Sandberg Esquenazi and Tal Svoray},
      year={2026},
      eprint={2512.15564},
      archivePrefix={arXiv},
      primaryClass={cs.CV},
      url={https://arxiv.org/abs/2512.15564}, 
}

@misc{yuan2024rrsisreferringremotesensing,
      title={RRSIS: Referring Remote Sensing Image Segmentation}, 
      author={Zhenghang Yuan and Lichao Mou and Yuansheng Hua and Xiao Xiang Zhu},
      year={2024},
      eprint={2306.08625},
      archivePrefix={arXiv},
      primaryClass={cs.CV},
      url={https://arxiv.org/abs/2306.08625}, 
}

@inproceedings{
wang2026disasterm3,
title={DisasterM3: A Remote Sensing Vision-Language Dataset for Disaster Damage Assessment and Response},
author={Junjue Wang and Weihao Xuan and Heli Qi and Zhihao Liu and Kunyi Liu and Yuhan Wu and Hongruixuan Chen and Jian Song and Junshi Xia and Zhuo Zheng and Naoto Yokoya},
booktitle={The Thirty-ninth Annual Conference on Neural Information Processing Systems Datasets and Benchmarks Track},
year={2026},
url={https://openreview.net/forum?id=sQO1ZEQGqX}
}

@article{sarlang1m,
  author={Wei, Yimin and Xiao, Aoran and Ren, Yexian and Zhu, Yuting and Chen, Hongruixuan and Xia, Junshi and Yokoya, Naoto},
  journal={IEEE Transactions on Geoscience and Remote Sensing}, 
  title={SARLANG-1M: A Benchmark for Vision–Language Modeling in SAR Image Understanding}, 
  year={2026},
  volume={64},
  pages={1-20},
  doi={10.1109/TGRS.2026.3652099}
  }

@InProceedings{geobenchvlm,
    author    = {Danish, Muhammad and Munir, Muhammad Akhtar and Shah, Syed Roshaan Ali and Kuckreja, Kartik and Khan, Fahad Shahbaz and Fraccaro, Paolo and Lacoste, Alexandre and Khan, Salman},
    title     = {GEOBench-VLM: Benchmarking Vision-Language Models for Geospatial Tasks},
    booktitle = {Proceedings of the IEEE/CVF International Conference on Computer Vision (ICCV)},
    month     = {October},
    year      = {2025},
    pages     = {7132-7142}
}

@article{rshrbench,
  title={A Benchmark for Ultra-High-Resolution Remote Sensing MLLMs},
  author={Dang, Yunkai and Zhu, Meiyi and Wang, Donghao and Zhang, Yizhuo and Yang, Jiacheng and Fan, Qi and Yang, Yuekun and Li, Wenbin and Miao, Feng and Gao, Yang},
  journal={arXiv preprint arXiv:2512.17319},
  year={2025}
}

@article{luo2026vlrsbench,
  title   = {VLRS-Bench: A Vision-Language Reasoning Benchmark for Remote Sensing},
  author  = {Luo, Zhiming and Wang, Di and Wang, Hebaixu and Guo, Haonan and Zhang, Jing and Du, Bo},
  journal = {arXiv preprint arXiv:2602.07045},
  year    = {2026}
}

@misc{jiang2026geoseg2,
      title={GeoSeg: Training-Free Reasoning-Driven Segmentation in Remote Sensing Imagery}, 
      author={Lifan Jiang and Yuhang Pei and oxi Wu and Yan Zhao and Tianrun Wu and Shulong Yu and Lihui Zhang and Deng Cai},
      year={2026},
      eprint={2603.03983},
      archivePrefix={arXiv},
      primaryClass={cs.CV},
      url={https://arxiv.org/abs/2603.03983}, 
}

@inproceedings{bert-score,
  title={BERTScore: Evaluating Text Generation with BERT},
  author={Tianyi Zhang* and Varsha Kishore* and Felix Wu* and Kilian Q. Weinberger and Yoav Artzi},
  booktitle={International Conference on Learning Representations},
  year={2020},
  url={https://openreview.net/forum?id=SkeHuCVFDr}
}

\newpage
\section*{Supplementary Material}
\appendix

\section{Trends analysis}
\label{app:mix_analysis}

\subsection{Setup, metrics, and covariates}
\label{app:mix_setup}
We analyze three training runs: an 8B model trained to 5k steps, a second 8B run with a different random seed trained to 3k steps, and a 2B model trained to 3k steps; all evaluated every 1k steps on the out-of-distribution (OOD) benchmark suite grouped into seven task domains. Unless stated otherwise, statistics are computed at the checkpoint from step 3k. 
These three runs share a single training mix, so Secs.~\ref{app:h1}--\ref{app:mv} are observational. We complement them with an interventional experiment (Sec.~\ref{app:intervention}): three additional 2B runs in which only the number of sources within selected domains is reduced.

Gain metric. Benchmarks use heterogeneous metrics and several base-model scores are near zero, our primary measure is the \emph{normalized gain}
\begin{equation}
\Delta_{\mathrm{norm}} = \frac{s_t - s_0}{1 - s_0},
\end{equation}
i.e., the fraction of the remaining headroom recovered by training; all metrics are mapped to $[0,1]$ beforehand. Absolute gains and $\log$-ratios (with $\varepsilon=0.005$) are used as sensitivity checks and never change the sign of any reported effect.

Training-mix covariates. Each OOD domain is matched to the training rows of the same task type and modality. Two covariates are derived per domain: $n_{\mathrm{smp}}$, the summed number of training samples per 1k steps, and $n_{\mathrm{src}}$, the number of unique underlying sources of labels (datasets may aggregate several sources). Sources shared across tasks are counted only once; detection additionally inherits the 13 segmentation sources because model produce bounding boxes for both tasks ($n_{\mathrm{src}}=19$); and the multi-modal VQA rows intersect heavily at the source level (DynamicVL, DisasterM3, SARLANG1, SARLANG2 contain SAR, GeoSeg-1M contain False-color images), yielding only 5 unique sources despite 16 nominal entries. The resulting counts are: Detection 19, Segmentation 13, RGB Caption 11, Multi-modal VQA 5, Ultra-high-res.\ VQA 3, Temporal VQA 2, Multi-view VQA 1. We notice, that $n_{\mathrm{smp}}$ and $n_{\mathrm{src}}$ are only mildly collinear ($\rho = 0.39$).

\begin{figure*}[t]
\centering
\includegraphics[width=\textwidth]{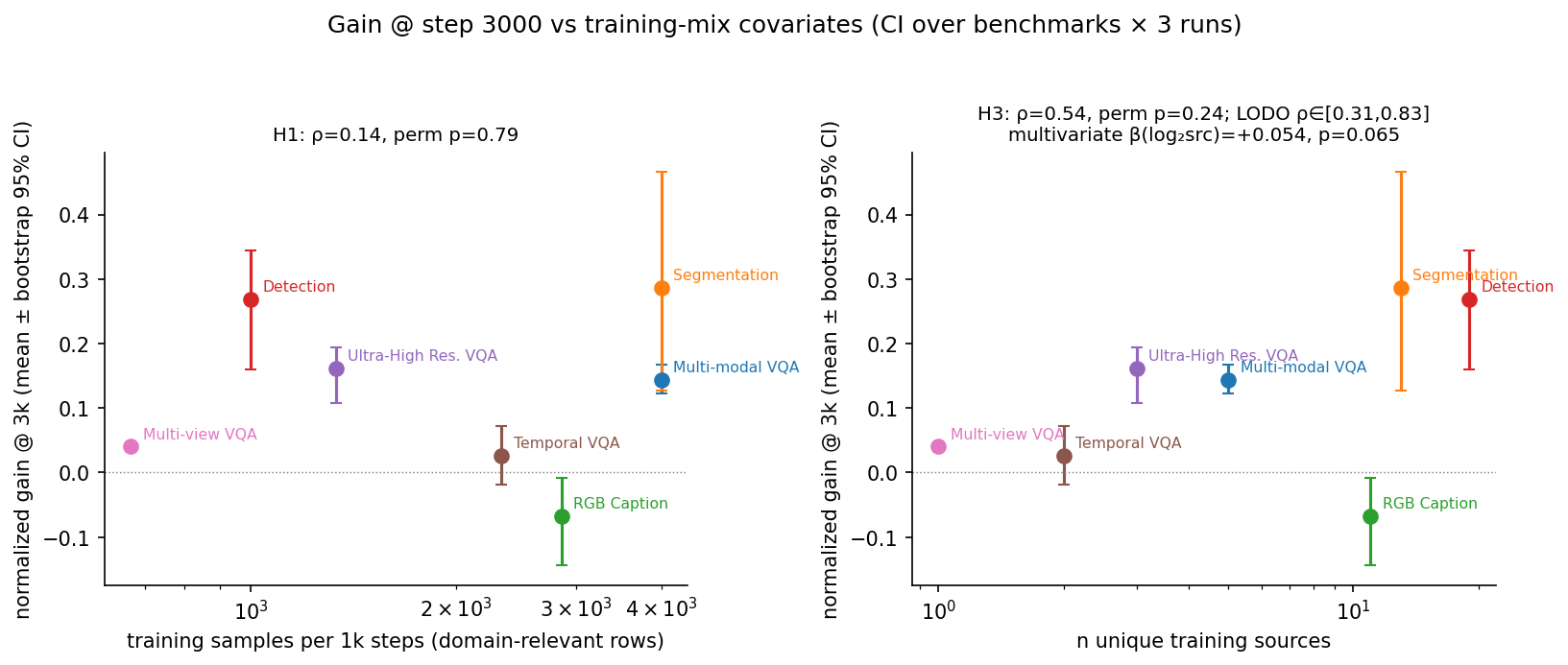}
\caption{Domain-level normalized gain at step 3000 (mean $\pm$ bootstrap 95\% CI over benchmarks $\times$ 3 runs) against the two training-mix covariates. \textbf{Left (H1):} no dependence on sampling volume. \textbf{Right (H3):} positive dependence on the number of unique sources; the visible off-trend point is RGB Caption, whose negative gain is explained by base-model saturation (Fig.~\ref{fig:base}) rather than lack of diversity.}
\label{fig:covariates}
\end{figure*}

Domains are not statistically independent: they share underlying sources, and the two 8B runs share initialization. We therefore use permutation tests for all univariate rank correlations, report leave-one-domain-out (LODO) ranges, cluster standard errors by domain in all regressions, and obtain confidence intervals for domain-level means by bootstrap over benchmark$\times$run observations within each domain. All visual error bars in Figs.~\ref{fig:covariates}--\ref{fig:seed_mv} are bootstrap 95\% CIs unless noted.

We further anayse effect of random seed, model size and base model performance. The hypothesis are: H2 and H1 -- behaviour is independent of seed, model size, and $n_{\text{smp}}$; H3 and H4 -- source diversity and base model performance affect behavior.

\begin{figure*}[t]
\centering
\includegraphics[width=\textwidth]{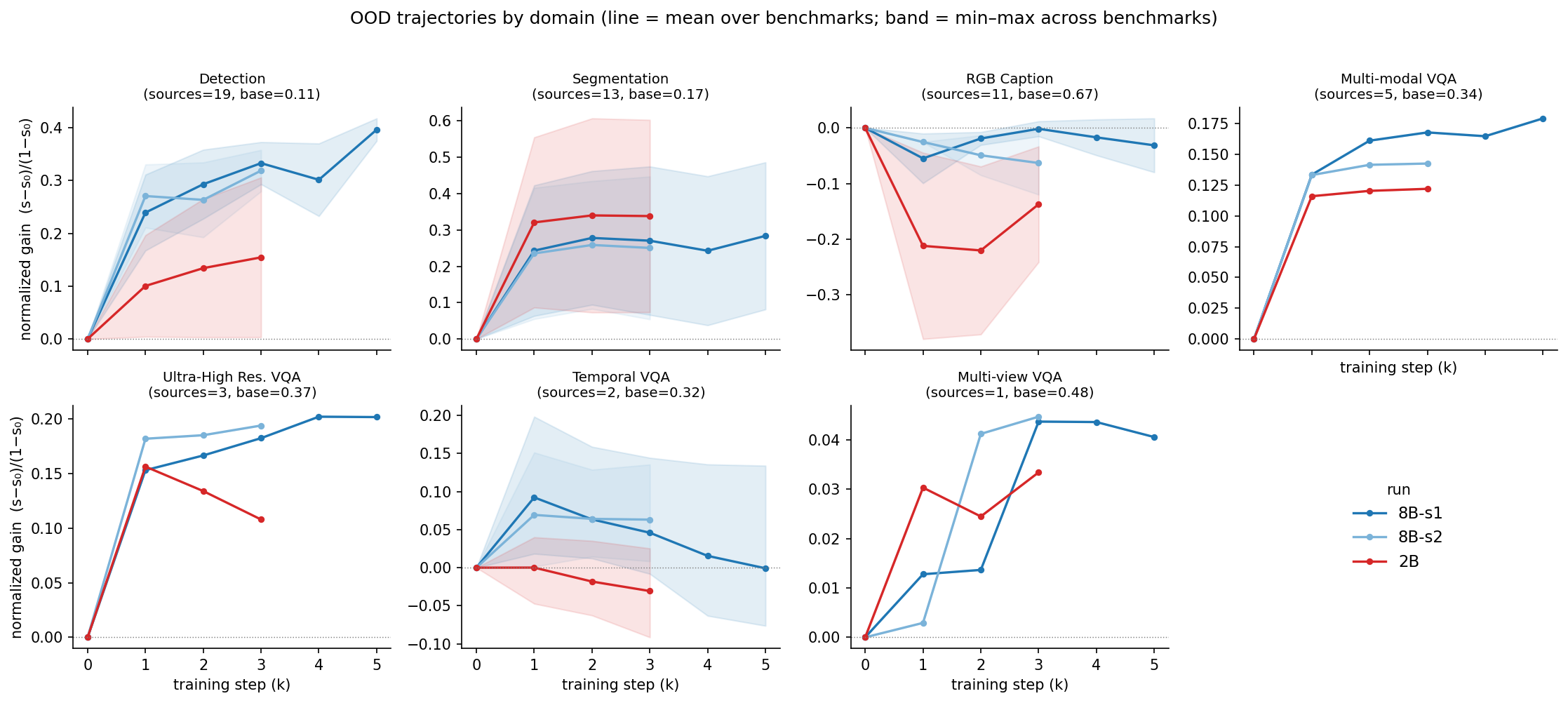}
\caption{OOD normalized-gain trajectories per task domain (panels ordered by decreasing $n_{\mathrm{src}}$). Lines are means over the benchmarks in each domain, bands are min--max ranges across benchmarks; colors denote the three runs. The three qualitative behaviours are visible: sustained improvement in source-rich domains (Detection, Segmentation, Multi-modal VQA), early peak followed by stagnation or decline in source-poor domains (Temporal VQA peaks at step 1k--2k in all three runs; Multi-view VQA gains remain $\lesssim 0.04$), and flat-to-negative movement where the base model is already saturated (RGB Caption, BERTScore base $\approx 0.88$).}
\label{fig:trajectories}
\end{figure*}

\subsection{H1: Gains are not driven by sampling volume}
\label{app:h1}
We find no evidence that $n_{\mathrm{smp}}$ predicts per-domain gains (Fig.~\ref{fig:covariates}, left). At the domain level ($n{=}7$), Spearman $\rho = 0.14$ (permutation $p = 0.79$; bootstrap 95\% CI $[-0.21, 0.36]$); Pearson $r = 0.10$ ($p = 0.82$). In the multivariate model of Sec.~\ref{app:mv}, which controls for source diversity and base performance, the coefficient on $\log_2 n_{\mathrm{smp}}$ is small and non-significant ($\beta = -0.015$ per SD, $p = 0.45$), and remains non-positive in all sensitivity specifications.

\subsection{H2: Seed and model scale do not change the behavior}
\label{app:h2}
Seed. Comparing the two 8B seeds pairwise over all 25 benchmarks and tasks (Fig.~\ref{fig:seed_mv}a), score differences are not significant at any shared checkpoint (Wilcoxon signed-rank $p = 0.11$, $0.43$, $0.31$ at steps 1k, 2k, 3k), and the mean absolute seed difference ($0.014$) is $7.3\times$ smaller than the mean absolute training gain ($0.10$). Per-benchmark normalized gains at 3k are nearly identical across seeds (Spearman $\rho = 0.965$, $p < 10^{-6}$), i.e., not only the magnitudes but the \emph{pattern} of gains is reproducible.

Model size. The 2B model reproduces the same gain structure as the 8B models on the OOD suite (Fig.~\ref{fig:seed_mv}b): per-benchmark gains correlate at $\rho = 0.77$ ($p = 0.003$) with the seed-averaged 8B gains. The 2B gains are marginally smaller in magnitude (median difference $-0.048$, Wilcoxon $p = 0.052$), driven mainly by detection and ultra-high-resolution VQA (Fig.~\ref{fig:trajectories}). Trajectory-shape classifications (sustained vs.\ peaked, Sec.~\ref{app:h3}) agree across seeds and across scales on 57\% of domains, with all disagreements occurring in low-amplitude domains where the classification is noise-limited. 

We conclude that neither seed nor scale qualitatively alters the mix-dependent behavior, and pool all three runs elsewhere in this analysis.

\subsection{H3: Gains and trajectory shape increase with source diversity}
\label{app:h3}
Overall correlation. The domain-level association between gain at 3k and $n_{\mathrm{src}}$ is positive and moderate (Fig.~\ref{fig:covariates}, right): Spearman $\rho = 0.54$ (permutation $p = 0.24$ at $n{=}7$; bootstrap 95\% CI $[0.39, 0.64]$; LODO range $[0.31, 0.83]$). The univariate test is underpowered at seven domains and is additionally confounded by base-model saturation: RGB Caption has 11 sources yet negative gains because its base score is already $\approx 0.67$ on average. Controlling for base performance, the partial rank correlation remains positive ($\rho = 0.26$), and in the multivariate model (Sec.~\ref{app:mv}) $\log_2 n_{\mathrm{src}}$ carries the largest positive standardized coefficient ($\beta = +0.054$ per SD, $p = 0.065$; $+0.068$ excluding detection; $+0.050$ under Huber robust regression) --- consistently positive across every specification we tried, suggesting significance, however base performance competes for the shared variance. Restricting the analysis to the two 8B runs to remove any model-scale effect strengthens the association: domain-level $\rho = 0.61$ (permutation $p = 0.17$; LODO $[0.37, 0.94]$), and the multivariate sources coefficient becomes significant ($\beta = +0.069$, $p = 0.034$).  The interventional ablation in Sec.~\ref{app:intervention} confirms the direction of this effect: it varies $n_{\mathrm{src}}$ while holding the base model and benchmarks fixed.

Trajectory behaviour. We classify each domain$\times$run trajectory using min--max-normalized scores: \emph{drop from peak} (peak minus final value) and \emph{monotonicity} (Spearman correlation of score with step). The source-poor domains exhibit the failure mode: Temporal VQA ($n_{\mathrm{src}}{=}2$) peaks at step 1/2k and declines in all three runs (mean drop $0.42$), and Multi-view VQA ($n_{\mathrm{src}}{=}1$) improves only marginally. Tests over the seven domains are directionally consistent but not yet significant (peak retention vs.\ $n_{\mathrm{src}}$: $\rho = 0.52$, permutation $p = 0.24$; sustained vs.\ peaked domains, one-sided Mann--Whitney $p = 0.57$), again limited by $n{=}7$ and by RGB Caption, whose decline is saturation-driven rather than diversity-driven. We therefore present the behaviour-type result (Fig.~\ref{fig:trajectories}) as a qualitative, seed- and scale-reproducible observation rather than a confirmatory finding. The decline in temporal VQA after the peak is yet to be explained by, e.g., effects of forgetting. 

\begin{figure}[t]
\centering
\includegraphics[width=.6\linewidth]{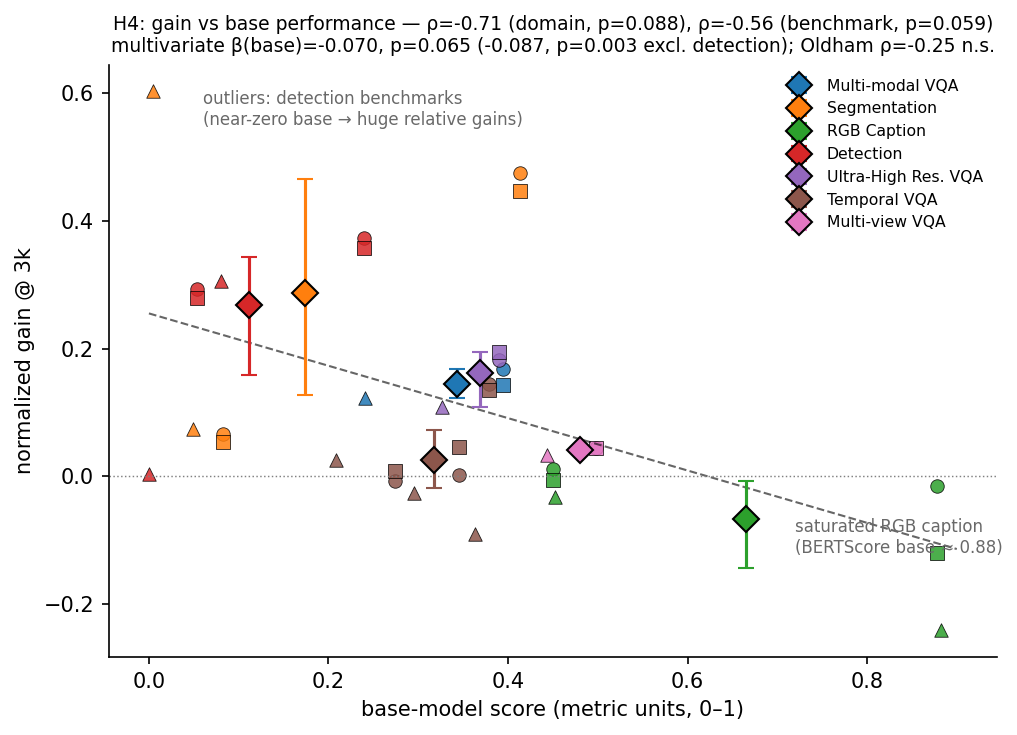}
\caption{Normalized gain at step 3000 against base-model score. Small markers: individual benchmarks per run (circle/square/triangle = 8B-s1/8B-s2/2B); large diamonds: domain means with bootstrap 95\% CIs; dashed line: OLS fit on benchmark means. Detection benchmarks (near-zero base) and RGB Caption (base $\approx 0.88$ BERTScore) anchor the two extremes of the headroom effect.}
\label{fig:base}
\end{figure}

\subsection{H4: Gains decrease with base-model performance}
\label{app:h4}
Gains correlate negatively with base performance (Fig.~\ref{fig:base}): Spearman $\rho = -0.71$ at the domain level (permutation $p = 0.088$) and $\rho = -0.56$ over the 12 OOD benchmarks ($p = 0.059$); in the multivariate model $\beta_{\mathrm{base}} = -0.070$ per SD ($p = 0.065$; $-0.087$, $p = 0.003$ excluding detection; $-0.069$ under Huber). However, the effect is not linear, it is rather clearest at the extremes: RGB Caption is saturated at base (BERTScore $\approx 0.88$; gains $\approx 0$ or negative in all runs), while detection starts near zero and improves up to $\times 7.6$ in relative terms. The conservative reading is that base-model headroom is a necessary condition for mix-driven gains.

\begin{figure*}[t]
\centering
\includegraphics[width=\textwidth]{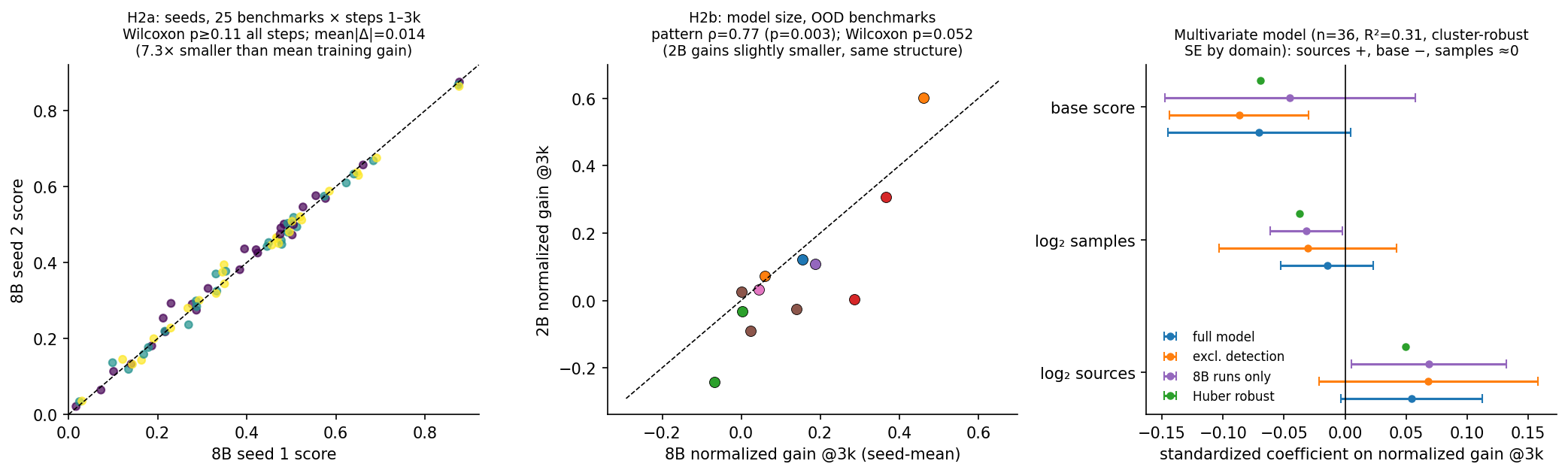}
\caption{\textbf{(a)} Seed comparison: 8B seed-1 vs.\ seed-2 scores over 25 benchmarks at steps 1k--3k (color = step); points lie on the identity line. \textbf{(b)} Scale comparison: 2B vs.\ 8B (seed-mean) normalized gains at 3k per OOD benchmark (colors = domains as in Fig.~\ref{fig:base}); the pattern is preserved with slightly attenuated 2B magnitudes. \textbf{(c)} Standardized coefficients of the multivariate model (Eq.~\ref{eq:mv}) with cluster-robust 95\% CIs, for the full model, excluding detection benchmarks, and under Huber robust regression: source diversity contributes positively, base performance negatively, and sampling volume contributes nothing, across all specifications.}
\label{fig:seed_mv}
\end{figure*}

\subsection{Multivariate model}
\label{app:mv}
To disentangle the three covariates we fit, at the benchmark level (12 OOD benchmarks $\times$ 3 runs, $n = 36$),
\begin{equation}
\Delta_{\mathrm{norm}}^{(3k)} = \beta_0 + \beta_1 \, z\!\left(\log_2 n_{\mathrm{src}}\right) + \beta_2 \, z\!\left(\log_2 n_{\mathrm{smp}}\right) + \beta_3 \, z\!\left(s_0\right) + \epsilon,
\label{eq:mv}
\end{equation}
with standard errors clustered by domain (7 clusters), where $z(\cdot)$ denotes standardization. Results are shown in Fig.~\ref{fig:seed_mv}c. The model explains $R^2 = 0.31$ of benchmark-level gain variance with only these three mix/base covariates. Source diversity and base headroom carry comparable, opposite-signed weights, while sampling volume is null --- the quantitative counterpart of H1, H3, and H4. Because $n_{\mathrm{src}}$ and $s_0$ are themselves anti-correlated in our mix (source-rich domains happen to have low-scoring bases), the two coefficients share variance and each is only marginally significant in the full model; their signs and magnitudes are nevertheless stable across outlier removal and robust estimation.

\begin{table}[t]
\centering
\setlength{\tabcolsep}{2pt}
\caption{Interventional ablation of $n_{\mathrm{src}}$ (2B model, 3k steps, fixed number of training samples and steps). The first row is the reference mixture; following rows reduce $n_{\mathrm{src}}$ in the domains: det/seg/cap/mm/uhr: detection, segmentation, RGB caption, multi-modal VQA, and ultra-high-resolution VQA. \textcolor{red}{Red} marks targeted intervention in $n_{\mathrm{src}}$ and corresponding benchmarks.}
\label{tab:nsrc_intervention}
\resizebox{\linewidth}{!}{%
\begin{tabular}{ccccccccccccc}
\toprule
    Mixture & ~ & ~ & GEO & ~ & ~ & GSeg & RSHR & UrB & VLRS & ~ & XLRS & ~ \\
    det/seg/cap/mm/uhr & cap & vqa & chdet & det & seg & seg & vqa & vqa & vqa & cap & det & vqa \\
\midrule
    19~/~13~/~11~/~5~/~3 & 85.4 & 33.4 & 27.7 & 36.2 & 12.0 & 60.5 & 30.5 & 46.2 & 22.9 & 43.4 & 3.3 & 39.9 \\
    ~\textcolor{red}{9}~/~~\textcolor{red}{9}~~/~11~/~5~/~3 & 85.8 & 41.8 & 32.6 & \textcolor{red}{31.8} & \textcolor{red}{11.4} & \textcolor{red}{58.3} & 32.4 & 47.6 & 24.7 & 43.4 & \textcolor{red}{3.0} & 40.4 \\
    19~/~13~/~11~/~\textcolor{red}{4}~/~\textcolor{red}{1} & 78.6 & \textcolor{red}{34.0} & 29.2 & 39.6 & 16.3 & 62.7 & 34.3 & 46.4 & 21.9 & 41.7 & 3.7 & \textcolor{red}{36.6} \\
    ~\textcolor{red}{17}~/~13~/~~\textcolor{red}{5}~~/~\textcolor{red}{3}~/~3 & \textcolor{red}{78.0} & \textcolor{red}{25.3} & 26.8 & \textcolor{red}{34.0} & 8.9 & 59.3 & 33.7 & 46.3 & 22.6 & \textcolor{red}{43.4} & \textcolor{red}{3.0} & 42.0 \\
\bottomrule
\end{tabular}%
}
\end{table}

\subsection{Interventional test of H3: ablating source diversity}
\label{app:intervention}
The observational analysis cannot fully separate source diversity from base-model performance, because the two are anti-correlated across domains in our mix (Sec.~\ref{app:mv}). Thus, we vary $n_{\mathrm{src}}$ directly. Starting from the original mixture, we train three additional 2B models for 3k steps, changing only the number of label sources within selected domains: (i) detection $19\!\to\!9$ and segmentation $13\!\to\!9$ via removal of GeoZero detection, DynamicVL and DisasterM3 segmentation; (ii) multi-modal VQA $5\!\to\!4$ and ultra-high-resolution VQA $3\!\to\!1$ via removal of GeoLLaVA, and GeoZero vqa; (iii) detection $19\!\to\!17$, RGB caption $11\!\to\!5$, and multi-modal VQA $5\!\to\!3$ via removal of SARLANG2, and all GeoZero.
The total number of training samples and optimizer steps are identical across runs. Every comparison is made within the same benchmark, therefore the base score $s_0$ is fixed, removing the $n_{\mathrm{src}}$--$s_0$ confound of the multivariate model.

Results are shown in Tab.~\ref{tab:nsrc_intervention}. Of the 11 benchmark scores targeted by a source reduction, 9 decrease, 1 is unchanged (XLRS-Bench caption in (iii)), and 1 increases (GEOBench-VLM VQA in (ii)); the mean change is $-2.6$ points on targeted benchmarks versus $+1.0$ on non-targeted ones due to freed model capacity for other domains. Removing sources lowers the corresponding scores, and we confirm the association between gains and $n_{\mathrm{src}}$ with $p = 0.04$. Combining this interventional evidence with the observational analysis above, yields $p = 0.008$.

\subsection{Outlier handling and limitations}
\label{app:limits}
Outliers. XLRS-Bench detection (relative gain up to $\times 7.6$; 2B base score exactly $0$) and GeoSeg-Bench2 (2B base $0.004$) would dominate any ratio-based analysis; however, the normalized-gain metric $\Delta_{\text{norm}}$ bounds their influence, and all results were re-checked excluding detection benchmarks and under Huber robust regression, without sign changes. 

The main limitations are: statistical power -- seven task domains cap the resolution of all domain-level univariate tests, and the behaviour-type association in particular should be treated as observational; the source-sharing dependence between domains, which we mitigate with clustered errors, permutation tests, and LODO; and the collinearity between source diversity and base headroom in this particular mix, which prevents a fully clean attribution between H3 and H4 from the observational runs alone. The interventional ablation (Sec.~\ref{app:intervention}) addresses this by holding $s_0$ fixed, using single runs for a model of smaller size, assuming H2. Finally, diversity require stricter measure than number of annotation sources $n_{\mathrm{src}}$.

\section{Detailed Multi-Task Reward}
For each sampled completion, we assign a scalar reward that combines a small format bonus with a task-specific correctness score:
\begin{equation}
  R = \lambda_{\mathrm{fmt}} F + \lambda_{\mathrm{task}} S,
  \label{eq:reward_general}
\end{equation}
where $F \in \{0,1\}$ indicates whether the output follows the required interface, $S \in [0,1]$ is the task-specific score. We use common~\cite{markov2026bgrto,shao2024deepseekmath} default values $\lambda_{\mathrm{fmt}}=0.1$ and $\lambda_{\mathrm{task}}=0.9$ during training.

For multiple-choice VQA, the model must answer using the strict \\ \texttt{<answer>...</answer>} format. The reward supports both letter-based options, such as \texttt{A}, \texttt{B}, and \texttt{C}, and text-based options, such as class names or true/false labels. Let $\hat{A}$ be the parsed set of predicted options, $A$ be the target answer set, and $\Omega$ be the set of valid options. The format term is one only if the answer can be parsed and all predicted options are valid:
\begin{equation}
  F^{\mathrm{mc}} =
  \mathbbm{1}\left[
  \hat{A} \neq \emptyset
  \;\wedge\;
  \hat{A} \subseteq \Omega
  \right].
  \label{eq:mc_format}
\end{equation}
The task score is set IoU between the predicted and target answer sets:
\begin{equation}
  S^{\mathrm{mc}} =
  \begin{cases}
    \dfrac{|\hat{A} \cap A|}{|\hat{A} \cup A|},
    & \text{if } F^{\mathrm{mc}}=1, \\[6pt]
    0,
    & \text{otherwise.}
  \end{cases}
  \label{eq:mc_score}
\end{equation}
This handles both single-answer and multi-answer multiple-choice questions.

For free-form VQA and captioning, the model must also produce a non-empty answer inside \texttt{<answer>...</answer>}. Let $\hat{a}$ be the extracted prediction and $a$ be the reference. The format term is
\begin{equation}
  F^{\mathrm{text}} =
  \mathbbm{1}\left[
  \hat{a} \text{ is non-empty and correctly enclosed in }
  \texttt{<answer>...</answer>}
  \right].
  \label{eq:text_format}
\end{equation}
When the answer is valid, we score it with Qwen3-Embedding-0.6B~\cite{zhang2025qwen3embedding}. The prediction and reference are embedded, last-token pooled, $\ell_2$-normalized, and compared with cosine similarity. We map the cosine score from $[-1,1]$ to $[0,1]$:
\begin{equation}
  S^{\mathrm{text}} =
  \begin{cases}
    \mathrm{clip}_{[0,1]}
    \left(
    \dfrac{
    1 + \cos\left(
    \phi(\hat{a}), \phi(a)
    \right)
    }{2}
    \right),
    & \text{if } F^{\mathrm{text}}=1, \\[8pt]
    0,
    & \text{otherwise,}
  \end{cases}
  \label{eq:text_score}
\end{equation}
where $\phi(\cdot)$ denotes the frozen Qwen3-Embedding-0.6B encoder. This gives a dense semantic reward for open-ended language outputs instead of relying on exact string matching.

For segmentation, the model must invoke the localization tool by producing a valid \texttt{<segmentation>...</segmentation>} call, containing a noun phrase and a list of bounding boxes for each object type. The language output is parsed into phrase-level prompts, which are passed to SAM to obtain semantic masks. The format term is the validity of this tool call:
\begin{equation}
  F^{\mathrm{seg}} =
  \mathbbm{1}\left[
  \text{the segmentation tool call is valid and yields SAM masks}
  \right].
  \label{eq:seg_format}
\end{equation}
Let $\hat{M}=\{\hat{m}^{k}\}_{k=1}^{P}$ be the predicted phrase-level masks and $M=\{m^{k}\}_{k=1}^{T}$ be the ordered target masks. If the tool call is valid and the number of predicted masks matches the number of target masks, we compute ordered semantic IoU:
\begin{equation}
  S^{\mathrm{seg}} =
  \begin{cases}
    \dfrac{1}{T}
    \displaystyle\sum_{k=1}^{T}
    \mathrm{IoU}\left(\hat{m}^{k}, m^{k}\right),
    & \text{if } F^{\mathrm{seg}}=1 \text{ and } P=T, \\[12pt]
    0,
    & \text{otherwise.}
  \end{cases}
  \label{eq:seg_score}
\end{equation}
Thus, each predicted semantic mask is compared with the target mask at the same position, and completions that produce the wrong number of masks receive zero segmentation score. If the task contains only one foreground semantic mask, this reduces to standard mask IoU:
\begin{equation}
  S^{\mathrm{seg}} =
  \mathrm{IoU}\left(\hat{m}, m\right).
  \label{eq:single_seg_score}
\end{equation}

For detection, the model uses the same localization interface and outputs the same structured JSON object inside \texttt{<segmentation>...</segmentation>}. The format term is one only when the JSON is valid, contains the required noun phrase field, and every object contains a valid box:
\begin{equation}
  F^{\mathrm{det}} =
  \mathbbm{1}\left[
  \text{the output is valid localization JSON with valid boxes}
  \right].
  \label{eq:det_format}
\end{equation}
Predicted boxes are converted to image pixels when they are produced in normalized or Qwen-style grid coordinates. Let $\hat{B}=\{\hat{b}^{p}\}_{p=1}^{P}$ be the predicted boxes and $B=\{b^{t}\}_{t=1}^{T}$ be the ground-truth boxes. We greedily match predicted and ground-truth boxes by IoU, using each box at most once, and average over the larger number of predicted and target boxes:
\begin{equation}
  S^{\mathrm{det}} =
  \begin{cases}
    \dfrac{1}{\max(P,T)}
    \displaystyle\sum_{(p,t)\in \mathcal{M}}
    \mathrm{IoU}\left(\hat{b}^{p}, b^{t}\right),
    & \text{if } F^{\mathrm{det}}=1, \\[12pt]
    0,
    & \text{otherwise,}
  \end{cases}
  \label{eq:det_score}
\end{equation}
where $\mathcal{M}$ is the set of matched prediction--target pairs. Across all tasks, the same reward form in Eq.~\eqref{eq:reward_general} is used, while the task-specific score $S$ changes according to the prompt type. This trains the language policy to either answer directly in text or invoke the localization tool when spatial output is required.

\section{Group Relative Tool Optimization}

We use Group Relative Tool Optimization (GRTO)~\cite{markov2026bgrto} to jointly adapt the language policy and the SAM3 localization tool. Standard tool-use reinforcement learning updates only the policy while keeping the tool fixed. In contrast, GRTO treats the VLM and tool as a coupled system: the VLM learns when and how to call the tool, while the tool is updated on the valid tool calls produced by the VLM.

Let $\pi_{\theta}$ denote the language policy and $\psi_{\omega}$ denote the parametrized tool. Both are initialized from pretrained reference models $\pi_{\theta_0}$ and $\psi_{\omega_0}$. The goal is to optimize the KL-regularized objective
\begin{equation}
  J_{\mathrm{KL\text{-}RL}}(\theta,\omega)
  =
  \mathbb{E}_{o\sim \pi_{\theta}}
  \left[
  R(o,\psi_{\omega})
  \right]
  -
  \beta\,
  D_{\mathrm{KL}}\!\left(\pi_{\theta}\,\|\,\pi_{\theta_0}\right),
  \label{eq:grto_klrl}
\end{equation}
where $o$ is a generated completion and $R$ is the task reward which depends on both policy and tool.

For the policy update, we use the GRPO objective. Given a group of $G$ completions $\{o_i\}_{i=1}^{G}$ sampled from the old policy $\pi_{\mathrm{old}}$, we compute rewards $\{R_i\}_{i=1}^{G}$, their group mean $\mu$, and standard deviation $\sigma$. The relative advantage of completion $o_i$ is
\begin{equation}
  A_i = \frac{R_i-\mu}{\sigma}.
  \label{eq:grto_advantage}
\end{equation}
The policy objective is
\begin{equation}
  J_{\mathrm{GRPO}}(\theta,\omega_{\mathrm{old}})
  =
  \frac{1}{G}
  \sum_{i=1}^{G}
  \sum_{t=1}^{|o_i|}
  \min\!\left(
  r_{i,t}A_i,
  \mathrm{clip}(r_{i,t},1-\epsilon,1+\epsilon)A_i
  \right)
  -
  \beta\,
  D_{\mathrm{KL}}\!\left(\pi_{\theta}\,\|\,\pi_{\theta_0}\right),
  \label{eq:grto_grpo}
\end{equation}
where
\begin{equation}
  r_{i,t}
  =
  \frac{
  \pi_{\theta}(o_{i,t}\mid o_{i,<t})
  }{
  \pi_{\mathrm{old}}(o_{i,t}\mid o_{i,<t})
  }.
  \label{eq:grto_ratio}
\end{equation}
The rewards in Eq.~\eqref{eq:grto_grpo} are computed using the current frozen copy of the tool, $\omega_{\mathrm{old}}$, so that the policy update follows the usual clipped GRPO form.

For the tool update, GRTO reuses the same sampled completions. Since the task reward is generally non-differentiable, the tool is optimized with a differentiable surrogate loss $\mathcal{L}(\psi_{\omega})$ on valid tool calls. The contribution of each rollout is weighted by the stopped-gradient policy likelihood ratio, yielding the GRTO objective
\begin{equation}
  J_{\mathrm{GRTO}}(\theta,\omega)
  =
  J_{\mathrm{GRPO}}(\theta,\omega_{\mathrm{old}})
  -
  \frac{1}{G}
  \sum_{i=1}^{G}
  \left(
  \prod_{t=1}^{|o_i|} r'_{i,t}
  \right)
  \mathcal{L}_i(\psi_{\omega}),
  \label{eq:grto_objective}
\end{equation}
where $r'_{i,t}=\mathrm{stopgrad}(r_{i,t})$. Thus, the language policy is updated by the GRPO reward signal, while the SAM3 tool is updated through the supervised auxiliary loss induced by the policy-generated tool calls.

\textbf{SAM3 auxiliary loss.}
We fine-tune the semantic head of SAM3 using the same BCE + soft-IoU objective as in B-GRTO~\cite{markov2026bgrto}. For each valid localization call, the VLM produces a noun phrase and bounding box prompts for SAM3. The SAM3 semantic head predicts a foreground logit mask $\hat{M}\in\mathbb{R}^{H\times W}$, which is converted into a soft probability mask
\begin{equation}
  S = \sigma(\hat{M}) \in [0,1]^{H\times W}.
  \label{eq:sam_soft_mask}
\end{equation}
Let $M\in\{0,1\}^{H\times W}$ be the binary ground-truth foreground mask. The auxiliary segmentation loss is
\begin{equation}
  \mathcal{L}_{\mathrm{sem}}
  =
  \mathcal{L}_{\mathrm{BCE}}
  +
  \mathcal{L}_{\mathrm{sIoU}}.
  \label{eq:sam_sem_loss}
\end{equation}
The binary cross-entropy term is
\begin{equation}
  \mathcal{L}_{\mathrm{BCE}}
  =
  -\frac{1}{HW}
  \sum_{u=1}^{H}
  \sum_{v=1}^{W}
  \left[
  M_{uv}\log S_{uv}
  +
  (1-M_{uv})\log(1-S_{uv})
  \right],
  \label{eq:sam_bce}
\end{equation}
and the soft-IoU term is
\begin{equation}
  \mathcal{L}_{\mathrm{sIoU}}
  =
  1
  -
  \frac{
  \sum_{u=1}^{H}
  \sum_{v=1}^{W}
  S_{uv}M_{uv}
  }{
  \sum_{u=1}^{H}
  \sum_{v=1}^{W}
  \left(S_{uv}+M_{uv}-S_{uv}M_{uv}\right)
  }.
  \label{eq:sam_siou}
\end{equation}
This loss is applied only to valid localization rollouts, i.e., completions that produce a non-empty, parseable SAM3 prompt with valid bounding boxes. Invalid localization calls and language-only completions do not contribute to the SAM3 auxiliary loss. In our implementation, mask logits are filtered by the VLM-provided bounding boxes before the loss is computed, ensuring that the tool update remains spatially tied to the regions selected by the VLM.

\section{SFT baseline}
\label{app:sft}
To isolate the effect of the RL objective, we train the 2B model with supervised fine-tuning (SFT) on the reference training mixture for the same 3k steps and compare it with GRTO under identical settings (Tab.~\ref{tab:sft}). SFT is better on both detection benchmarks and on the two high-resolution VQA benchmarks (RSHR, XLRS), but GRTO is better on the remaining 8 of 12 OOD metrics. The largest gaps are in segmentation, where SFT nearly collapses (GEO: $1.3$ vs.\ $12.0$; GeoSeg: $30.3$ vs.\ $60.5$), and in change detection ($18.5$ vs.\ $27.7$). SFT therefore improves individual tasks, but generalizes worse overall than GRTO.

\begin{table}[t]
\centering
\setlength{\tabcolsep}{2pt}
\caption{SFT vs.\ GRTO on the OOD benchmarks. Both use the 2B model, the same training mixture, and 3k steps. Column-wise maxima are shown in bold.}
\label{tab:sft}
\resizebox{\linewidth}{!}{%
\begin{tabular}{lcccccccccccc}
\toprule
    ~ & ~ & ~ & GEO & ~ & ~ & GSeg & RSHR & UrB & VLRS & ~ & XLRS & ~ \\
    Method & cap & vqa & chdet & det & seg & seg & vqa & vqa & vqa & cap & det & vqa \\
\midrule
    SFT  & 84.7 & 28.5 & 18.5 & \textbf{41.2} & 1.3 & 30.3 & \textbf{37.3} & 39.6 & 21.6 & 36.5 & \textbf{6.8} & \textbf{43.1} \\
    GRTO & \textbf{85.4} & \textbf{33.4} & \textbf{27.7} & 36.2 & \textbf{12.0} & \textbf{60.5} & 30.5 & \textbf{46.2} & \textbf{22.9} & \textbf{43.4} & 3.3 & 39.9 \\
\bottomrule
\end{tabular}%
}
\end{table}

\section{Training data mix}

We gather varied set of tasks and modalities: GeoZero corpus~\cite{wang2026geozero}, consisting of Hard And Instruct parts, for generic RGB; GeoLLaVA~\cite{wang2026geollavak} for high-res; GeoSeg-1M~\cite{ni2025unigeoseg}, LaSeRS~\cite{xin2025segearthr2}, EarthReason~\cite{li2025earthreason} for generic and false-colour segmentation; DynamicVL~\cite{xuan2026dynamicvl}, DisasterM3~\cite{wang2026disasterm3} for temporal tasks; SARLANG-1M~\cite{sarlang1m} for SAR imagery. SARLANG-1M contains captions and free-form VQA from two sources: text produced from pre-annotated bounding boxes, and text produced from co-registred RGB images -- we refer to them as SARLANG1 and SARLANG2. We also construct multi-view instruction data from CVG-Text~\cite{ye2025icrossviewgeolocalizationnatural}, a cross-view geo-localization corpus of co-registered ground-level panoramas, ground-view captions, and satellite images: panorama captions are used directly for captioning, and we deterministically generate two MC-VQA item types — city recognition from a panorama–satellite pair, and cross-view matching, where the model must select which of four satellite views (with same-city hard negatives) depicts the panorama's location. A total raw pool of collected training data consists of 2.3M image-prompt-answer pairs. We randomly select a 80k subset to ensure all samples are unique during training. To balance the tasks and sources we utilize the following mix:
\begin{itemize}
    \item \emph{Multi-choice VQA}: GeoLLaVA, GeoZero Hard, GeoZero Instruct, DynamicVL, DisasterM3, VQA constructed from CVG-text -- each 3333 samples;
    \item \emph{Free-form VQA}: GeoZero Hard, GeoZero Instruct, SARLANG1, SARLANG2, DynamicVL, DisasterM3 -- each 2500 samples;
    \item \emph{Caption}: GeoZero Hard, GeoZero Instruct, SARLANG1, SARLANG2, DynamicVL, DisasterM3, CVG-text -- each 2857 samples;
    \item \emph{Segmentation}: GeoSeg-1M, LaSeRs, EarthReason, DynamicVL, DisasterM3 -- each 4000 samples;
    \item \emph{Detection}: GeoZero Hard, GeoZero Instruct -- each 2500 samples.
\end{itemize}

Within each batch we further ensure the following composition of tasks: 25\% captioning, 25\% multi-choice vqa, 25\% segmentation, 18.75\% free-form vqa, 6.25\% detection.

We ensure training data doesn't share the same images with any testset. We cannot ensure spatial separation as benchmarks do not provide coordinates for the images. We define OOD primarily at the task/annotation level; this reflects realistic Earth-observation deployment: satellite coverage is inherently shared and geographically finite, whereas the key source of diversity is how the same visual world is queried, annotated, and reasoned over. We therefore do not enforce artificial geographic isolation.

\section{Implementation details} 
We resize images to have a maximum side length of 4030; if more than 3, the maximum length is 448. For all samples, we provide the images first, followed by the prompt, abiding by each dataset's evaluation protocol. The maximum context length is fixed to 8192. Segmentation tool (SAM3) is applied to the last RGB image of the input.

\section{Prompts Used}
During training, we replace dataset-specific prompts with generic task templates to avoid artificially inflating the model's performance. In particular, we do not want the model to exploit benchmark-specific phrasing as a shortcut for deciding how to answer. However, some datasets contain prompts that become ambiguous in a multi-task setting. For example, certain localization datasets provide only a description of the target object, without explicitly stating whether the object should be segmented or detected. In such cases, we minimally disambiguate the instruction by adding a task cue, such as prepending ``Segment the ...''. This reflects the expected deployment setting, where users specify the task they want the model to solve rather than providing only an underspecified statement.

\textbf{Main MLRS prompt.}
In addition to the task-specific instruction, we prepend a default system instruction to every prompt seen by the model during both training and inference. This instruction defines the two allowed output modes: a textual answer for language-only tasks and a segmentation tool call for localization tasks, along with their corresponding output formats. Using the same instruction across all settings provides a stable multi-task interface, making the expected behavior explicit and allowing reinforcement learning to focus on improving task performance rather than discovering the output protocol from scratch.
 
 \begin{tcolorbox}[
    breakable,
    colback=gray!5,
    colframe=gray!40,
    boxrule=0.4pt,
    sharp corners,
    fontupper=\small\ttfamily,
]
You are solving visual reasoning tasks.

You may respond in one of two final formats:

1. If the task requires segmenting/localizing objects in the image, call the segmentation tool by outputting:

\texttt{<segmentation>\{"noun phrase": "...", "objects": [\{"bbox": [x1, y1, x2, y2]\}]\}</segmentation>}

The segmentation JSON must contain:

- "noun phrase": exactly one referential noun phrase identifying the target.

- "objects": a list of objects, each with one "bbox" field.

- Coordinates should use the 0-999 image coordinate grid.
\\

If the task includes several images, segmentation tool uses the last rgb image.
\\

2. If the task asks for a textual answer and does not require segmentation, output:

\texttt{<answer>your textual answer</answer>}
\\

Use \texttt{<think>...</think>} before the final tag when reasoning is useful.

Do not use \texttt{<answer>} for segmentation. Do not use \texttt{<segmentation>} for non-segmentation answers.
\end{tcolorbox}

\textbf{G-Eval prompt for captioning.}
 \begin{tcolorbox}[
    breakable,
    colback=gray!5,
    colframe=gray!40,
    boxrule=0.4pt,
    sharp corners,
    fontupper=\small\ttfamily,
]
You are an expert judge evaluating satellite image captions. Your task is to compare the Predicted caption against the Ground Truth (GT) and assign a score based on object accuracy, counting, and hallucinations.

Evaluation Steps:

1. Analyze the Ground Truth for core objects and counts.

2. Check the Prediction for "Imaginary Objects" (Hallucinations) not present in the GT.

3. Verify if object counts and spatial relationships match the GT.

4. Assign a strict score from 1-5 using the rubric below.
\\

Scoring Rubric:

1 (Critical Failure): Major hallucination (imaginary objects) or completely wrong scene classification.

2 (Poor): Correct scene type, but severe errors in object counting or wrong object attributes.

3 (Fair): Captures the main gist, but has minor hallucinations or noticeable counting errors.

4 (Good): Accurate objects and counts, with only very minor semantic differences or missing fine details.

5 (Perfect): Exact match in object types, counts, and spatial layout with no hallucinations.
\\

Input:

Ground Truth: \{gt\}

Predicted: \{pred\}

Score:
\end{tcolorbox}

Where we insert the ground truth and provided captions in place of \{gt\}, and \{pred\} respectively.

\textbf{G-Eval prompt for free-form VQA.}
 \begin{tcolorbox}[
    breakable,
    colback=gray!5,
    colframe=gray!40,
    boxrule=0.4pt,
    sharp corners,
    fontupper=\small\ttfamily,
]
You are an expert judge evaluating Visual Question Answering (VQA) outputs. Your task is to compare the Predicted Answer against the Ground Truth (GT) Answer for the given Question and assign a score based on factual correctness, completeness, and hallucinations.

Evaluation Steps:

1. Analyze the Question to understand what information is required.

2. Examine the Ground Truth Answer for key facts, values, and constraints.

3. Check the Predicted Answer for hallucinations (information not supported by the GT).

4. Verify correctness, precision, and completeness of the Predicted Answer.

5. Assign a strict score from 1-5 using the rubric below.
\\

Scoring Rubric:

1 (Critical Failure): Incorrect answer or major hallucination; does not address the question.

2 (Poor): Partially related but mostly incorrect; major factual errors or missing key elements.

3 (Fair): Captures the general idea but contains minor errors, ambiguity, or incomplete details.

4 (Good): Mostly correct and complete; only very minor inaccuracies or omissions.

5 (Perfect): Exact match with the Ground Truth; fully correct, precise, and no hallucinations.
\\

Input:

Question: \{q\}

Ground Truth Answer: \{gt\}

Predicted Answer: \{pred\}

Score:
\end{tcolorbox}

Where we insert the question, ground truth, and provided answer in place of \{q\}, \{gt\}, and \{pred\} respectively.

\section{Per-sub-task results}
Several benchmarks provide fine split on reasoning and sub-tasks dimensions, therefore we report according performance of our base model and MLRS model, see Tables~\ref{tab:pt_xlrs}, \ref{tab:pt_rshr}, \ref{tab:pt_vlrs}, \ref{tab:pt_ur}, \ref{tab:pt_lasers}, \ref{tab:pt_gsb}.

\begin{table*}[ht!]
\footnotesize
\caption{
XLRS-Bench~\cite{xlrsbench} multi-choice VQA. Left-to-right order: OC=Overall counting, RC=Regional counting, OLUC=Overall Land use classification, RLUC=Regional Land use classification, OCC=Object classification, OCL=Object color, OMS=Object motion state, OSR=Object spatial relationship, AD=Anomaly Detection and Interpretation, ECR=Environmental condition reasoning, RP=Route planning, RCCD=Counting with changing detection, CCR=Counting with complex reasoning.  'Avg.' represents the average accuracy across sub-tasks.
}
\label{tab:pt_xlrs}
\centering
\resizebox{\textwidth}{!}{%
\begin{tabular}{l|cccccccc|ccccc|c}
\toprule
\multicolumn{1}{c}{\textbf{Method}} & \multicolumn{8}{c}{\textbf{Perception}}  & \multicolumn{5}{c}{\textbf{Reasoning}} &  \\  \midrule
\textbf{Sub-tasks (L-3)}  &\textbf{OC} & \textbf{RC} & \textbf{OLUC} & \textbf{RLUC}  & \textbf{OCC} & \textbf{OCL} & \textbf{OMS} & \textbf{OSR} & \textbf{AD}  & \textbf{ECR} & \textbf{RP} & \textbf{RCCD} & \textbf{CCR} & \textbf{Avg.} \\ \midrule
\multicolumn{15}{l}{\textit{\textcolor{gray}{Remote Sensing MLLMs}}} \\
GeoChat & 16.7 & 29.0 & 2.0 & 23.0 & 21.1 & 16.8 & 35.0 & 24.2 & 33.0 & 43.0 & 10.0 & - & 21.0 & 22.9\\
GeoLLaVA-8K & 26.7 & 38.0 & {49.0} & 69.0 & 41.6 & 31.6 & 65.0 & 35.0 & 67.0 & 78.0 & {66.0} & {50.0} & {52.0} & 51.5 \\
\midrule
\multicolumn{15}{l}{\textit{\textcolor{gray}{Closed-source MLLMs}}} \\ 
GPT-4o & 25.0 & 32.0 & 15.0 & 66.0 & 9.5 & 11.3 & 11.7 & 24.6 & 73.0 & 73.0 & 35.0 & 20.0 & 25.0 & 32.4\\
GPT-4o-mini  & 23.3 & 25.0 & 19.0 & 59.5 & 40.9 & 31.0 & 65.0 & 23.6 & 71.0 & 71.0 & 29.0 & 6.7 & 30.0 & 38.1\\
Claude 3.7 Sonnet & 27.6 & 22.7 & 17.4 & 68.4 & 30.5 & 29.9 & 63.6 & 27.6 & 64.8 & 78.4 & 34.5 & 27.8 & 32.6 & 40.5\\
Gemini 2.0 Flash & {41.7} & 45.0 & 38.0 & 73.5 & 34.6 & 27.6 & 61.7 & 32.0 & 73.0 & 82.0 & 43.0 & 30.0 & 51.0 & 48.7\\
\midrule
\multicolumn{15}{l}{\textit{\textcolor{gray}{Open-source MLLMs}}} \\
InternLM-XComposer-2.5 & 21.7 & 42.0 & 7.0 & 68.0 & 31.8 & 27.8 & 6.7 & 26.0 & 72.0 & 81.0 & 41.0 & 36.7 & 47.0 & 39.1\\
LLaVA-Next & 26.7 & 40.0 & 5.0 & 67.0 & 28.8 & 32.8 & {66.7} & 30.0 & 69.0 & 78.0 & 27.0 & 35.0 & 36.0 & 41.7\\
LLaVA-OneVision-7B & 25.0 & 38.0 & 2.0 & 69.5 & 35.9 & 35.3 & 65.0 & 25.2 & 76.0 & {83.0} & 24.0 & 43.3 & 36.0 & 42.9\\
InternVL3-8B & 40.0 & 39.0 & 10.0 & 71.5 & {44.5} & 30.8 & 65.0 & 25.2 & {77.0} & 82.0 & 36.0 & 21.7 & 50.0 & 45.6\\
Qwen2-VL-7B & 26.7 & 40.0 & 11.0 & 73.0 & 35.9 & 34.6 & 61.7 & 31.8 & 70.0 & 81.0 & 35.0 & 46.7 & 48.0 & 45.8\\
LLaVA-OneVision-72B & 33.3 & 38.0 & 15.0 & 72.5 & 36.3 & 36.3 & {66.7} & 35.6 & 74.0 & {83.0} & 28.0 & 36.7 & 43.0 & 46.0\\
InternVL2.5-8B & 38.3 & 37.0 & 10.0 & 77.0 & 33.4 & 35.5 & 65.0 & 21.6 & 73.0 & {83.0} & 34.0 & {50.0} & 43.0 & 46.2\\
Qwen2.5-VL-7B & 33.3 & 40.0 & 31.0 & 77.0 & 40.6 & 40.5 & {66.7} & {36.2} & 68.0 & 72.0 & 27.0 & 38.3 & 45.0 & 47.4\\
InternVL3-78B & 23.3 & {49.0} & 33.0 & 74.0 & 42.5 & 37.4 & {66.7} & 30.0 & 76.0 & 81.0 & 40.0 & 45.0 & 42.0 & 49.2\\
Qwen3-VL-8B & 21.7 & 50.0 & 26.0 & {81.5} & 46.6 & 43.1 & 66.7 & 30.4 & 74.0 & 79.0 & 37.0 & 43.3 & {51.0} & 50.0 \\
Qwen2.5-VL-72B & 33.3 & 47.0 & 39.0 & {80.0} & 45.3 & {42.1} & 65.0 & 34.0 & 71.0 & 74.0 & 37.0 & 43.3 & 42.0 & 50.2\\
Qwen3-VL-235B-A22B & {61.7} & {82.5} &  38.6 & 36.7 & 44.0 & 39.0 & 38.9 & {49.0} & 73.0 &  82.0 & 37.8 & 33.3 & 48.0 & 51.1 \\
\midrule
\textbf{Base} & 21.7 & 20.0 & 48.0 & 50.5 & 30.1 & 33.1 & 63.3 & 23.4 & 54.0 & 71.0 & 38.0 & 23.3 & 31.0 & 39.0 \\
\textbf{MLRS} & 38.3 & 39.0 & 47.0 & 78.0 & 35.8 & 33.5 & {66.7} & 31.8 & 70.0 & 80.0 & 52.0 & 43.3 & {52.0} & {51.3} \\
 \bottomrule
\end{tabular}%
}
\end{table*}

\begin{table*}[ht!]
\footnotesize
\vspace{-0.2cm}
\caption{RSHR-Bench~\cite{rshrbench} multi-choice VQA. Left-to-right order:
\emph{Perception}—{COL}=Color Detection, {SHP}=Shape Recognition, {DET}=Detection, {OC}=Object Classification, {REL}=Object Spatial Relationship, {OGD}=Object Grounding, {RG}=Regional Grounding, {OCN}=Object Counting, {RCN}=Regional Counting, {Avg.}=Perception average;
\emph{Reasoning}—{AD}=Anomaly (single-turn), {FP}=Future Prediction (multi-image), {MRJC}=Multi-region Joint Contrast (multi-image), {MRJCS}=Multi-region Joint Contrast (single-image, multi-box), {OSJ}=Object State Judgment (single-turn), {Avg.}=Reasoning average;
\emph{Multi-turn}—{MAD}=Anomaly, {MTFP}=Future Prediction, MOSJ=Object State Judgment, Avg.=Multi-turn average. Column-wise maxima are shown in bold.
}
\label{tab:pt_rshr}
\centering
\resizebox{\textwidth}{!}{
\begin{tabular}{l|ccccccccc|c|ccccc|c|ccc|c}
\toprule
\multirow{2}{*}{\textbf{Model}} &
\multicolumn{9}{c}{\textbf{Perception}} &
\multirow{2}{*}{\textbf{Avg.}} &
\multicolumn{5}{c}{\textbf{Reasoning}} &
\multirow{2}{*}{\textbf{Avg.}} &
\multicolumn{3}{c}{\textbf{Multi-turn}} &
\multirow{2}{*}{\shortstack{\textbf{Avg.}}}\\
\cmidrule(lr){2-10}\cmidrule(lr){12-16}\cmidrule(lr){18-20}
& \textbf{COL} & \textbf{SHP} & \textbf{DET} & \textbf{OC} & \textbf{REL} & \textbf{OGD} & \textbf{RG} & \textbf{OCN} & \textbf{RCN}
&
& \textbf{AD} & \textbf{FP} & \textbf{MRJC} & \textbf{MRJCS} & \textbf{OSJ}
&
& \textbf{MAD} & \textbf{MTFP} & \textbf{MOSJ}
&
\\
\midrule 
\multicolumn{21}{l}{\textit{\textcolor{gray}{Remote Sensing VLMs}}} \\
EarthDial & 41.0 & 22.0 & 21.0 & 30.0 & 32.5 & 30.5 & 27.1 & 18.0 & 31.0 & 28.1 & 42.0 & 30.0 & 29.5 & 32.0 & {52.0} & 37.1 & 56.7 & 60.0 & 73.5 & 63.4 \\
GeoChat & 32.5 & 22.0 & 24.0 & 29.5 & {40.0} & 25.0 & 22.9 & 22.5 & 29.0 & 25.9 & 30.0 & 24.0 & 25.5 & 30.0 & 32.0 & 28.3 & 48.3 & 46.0 & 62.9 & 52.4 \\
GeoLLaVA-8K & 25.0 & 24.0 & 25.0 & 25.0 & 25.0 & 25.0 & 21.4 & 25.0 & 25.0 & 24.5 & 24.0 & 0.0 & 0.0 & 34.0 & 22.0 & 16.0 & 25.0 & 24.7 & 47.7 & 32.5 \\
VHM & 25.5 & 25.0 & 26.0 & 26.5 & \textbf{55.0} & 25.0 & 22.9 & 25.0 & 25.0 & 25.7 & 26.0 & 24.0 & 26.5 & 34.0 & 28.0 & 27.7 & 45.0 & 53.3 & 46.2 & 48.2 \\
\midrule
\multicolumn{21}{l}{\textit{\textcolor{gray}{Open-source VLMs}}} \\
InternVL2.5-8B & 25.5 & 22.0 & 26.0 & 26.0 & 22.5 & 24.5 & 30.0 & 22.5 & 20.0 & 24.3 & 26.0 & 20.0 & 22.5 & 34.0 & 20.0 & 24.5 & 25.0 & 28.7 & 35.6 & 29.8 \\
InternVL3.5-8B & 21.5 & 28.0 & 18.0 & 21.5 & 29.0 & 28.5 & 30.0 & \textbf{26.5} & 25.0 & 25.3 & 20.0 & 16.0 & 29.0 & 34.0 & 26.0 & 25.0 & 30.0 & 22.7 & 40.2 & 31.0 \\
MiniCPM2\_6 & 21.5 & 28.0 & \textbf{30.0} & 24.0 & 19.5 & 29.5 & 34.3 & 22.0 & 29.0 & 27.4 & 26.0 & 30.0 & {35.0} & 32.0 & 30.0 & 30.6 & 26.7 & 23.3 & 31.1 & 27.0 \\
Phi-3.5-Vision & 25.0 & 24.0 & 25.0 & 25.0 & 23.5 & 25.0 & 22.9 & 25.0 & 25.0 & 24.5 & 24.0 & 22.0 & 23.5 & {30.0} & 22.0 & 24.3 & 28.3 & 24.7 & 47.0 & 33.3 \\
Qwen2.5-VL-7B & {29.5} & 25.0 & 22.0 & 28.0 & 25.0 & 24.5 & 24.3 & \textbf{26.5} & 22.0 & 25.2 & 26.0 & 28.0 & 25.0 & 10.0 & 20.0 & 21.8 & 21.7 & 24.0 & 10.6 & 18.8 \\
Deepseek-VL & 22.5 & 22.0 & 21.0 & 25.0 & 20.5 & 26.0 & 28.6 & 20.5 & 22.0 & 23.1 & 22.0 & 28.0 & {50.0} & 32.0 & 20.0 & 30.4 & 20.0 & 23.3 & 33.3 & 25.5 \\
VILA-HD & 40.0 & 22.0 & 22.0 & 37.0 & 35.5 & 26.0 & 21.4 & 24.5 & 24.0 & 28.0 & {58.0} & 30.0 & \textbf{55.0} & 32.0 & {58.0} & 46.6 & 65.0 & 57.3 & 57.6 & 60.0 \\
\midrule
\multicolumn{21}{l}{\textit{\textcolor{gray}{Closed-source VLMs}}} \\
GPT5 & 29.0 & 10.0 & 23.0 & 23.0 & 37.0 & 24.5 & 31.4 & 20.0 & 23.0 & 24.5 & \textbf{74.0} & \textbf{58.0} & 35.0 & 34.0 & \textbf{66.0} & \textbf{53.4} & \textbf{78.3} & \textbf{73.3} & \textbf{86.4} & \textbf{79.3} \\
GPT-4o & 49.5 & 23.0 & 15.0 & 35.5 & 30.5 & 28.0 & 27.1 & 22.5 & 41.0 & 30.2 & {68.0} & 56.0 & 30.5 & 32.0 & 64.0 & 50.1 & 70.0 & 72.0 & 84.1 & 75.4 \\
GPT-4o-mini & 41.5 & 16.0 & 29.0 & 31.5 & 31.5 & 32.0 & 28.6 & 19.5 & 32.0 & 29.1 & {54.0} & {54.0} & 31.5 & \textbf{48.0} & {54.0} & 48.3 & \textbf{78.3} & 68.0 & 75.0 & 73.8 \\
Gemini-2.5-pro & \textbf{55.0} & 18.0 & 31.0 & \textbf{40.0} & 41.5 & 32.5 & \textbf{45.7} & 25.0 & 25.0 & \textbf{34.9} & {66.0} & 32.0 & 41.5 & 38.0 & 50.0 & 45.5 & 56.7 & 60.0 & 57.6 & 58.1 \\
\midrule
\textbf{Base } & 50.5 & \textbf{35.0} & 27.0 & 36.0 & 26.0 & 22.0 & 22.9 & 12.0 & \textbf{43.0} & 30.5 & 60.0 & 14.0 & 10.0 & 12.0 & 58.0 & 30.8 & 41.3 & 66.7 & 50.0 & 52.7 \\
\textbf{Best eval (1k step) } & 46.5 & 24.0 & 27.0 & 34.0 & 30.5 & \textbf{37.5} & 35.7 & 24.5 & 41.0 & 33.5 & 56.0 & 24.0 & 30.0 & 38.0 & 48.0 & 37.0 & 40.0 & 66.7 & 50.0 & 35.7\\
\textbf{MLRS (5k step) } & 48.0 & 30.0 & 9.0 & 30.5 & 16.5 & 24.5 & 10.0 & 21.0 & 38.0 & 25.3 & 44.0 & 10.0 & 30.0 & 30.0 & 38.0 & 30.4 & 30.0 & 55.3 & 36.7 & 40.7 \\
\bottomrule
\end{tabular}
}
\end{table*}

\begin{table*}[ht!]
\centering
\caption{VLRS-Bench~\cite{luo2026vlrsbench} multi-choice VQA. Left-to-right order: CR=Causal Reasoning, CFR=Counterfactual Reasoning, SIR=Semantic Integration Reasoning, MIR=Mechanistic Interaction Reasoning, ST-CFR=Spatiotemporal Counterfactual Reasoning, ST-CCR=Spatiotemporal Causal-Chain Reasoning, SR-ER=Spatiotemporal Evolution Reasoning, ST-CR=Spatiotemporal Consistency Reasoning, PR=Planning Reasoning, ER=Evaluation Reasoning, ST-CS-PR=Spatiotemporal Category-State Prediction Reasoning, ST-M-PR=Spatiotemporal Morphological Prediction Reasoning, ST-SU-PR=Spatiotemporal Scenario Uncertainty Prediction Reasoning, ST-SQ-PR=Spatiotemporal Sequence Prediction Reasoning. Column-wise maxima are shown in bold.}
\label{tab:pt_vlrs}
\resizebox{\textwidth}{!}{
\begin{tabular}{l | cccccccc | cc | cccc | c}
\toprule
\multirow{2}{*}{\textbf{Models}} & \multicolumn{8}{c}{\textbf{Cognition}} & \multicolumn{2}{c}{\textbf{Decision}} & \multicolumn{4}{c}{\textbf{Prediction}} & \multirow{2}{*}{\textbf{Avg. Score.}} \\
\cmidrule(lr){2-9} \cmidrule(lr){10-11} \cmidrule(lr){12-15}
& \textbf{CR} & \textbf{CFR} & \textbf{SIR} & \textbf{MIR} & \textbf{ST-CFR} & \textbf{ST-CCR} & \textbf{ST-ER} & \textbf{ST-CR} & \textbf{PR} & \textbf{ER} & \textbf{ST-CS-PR} & \textbf{ST-M-PR} & \textbf{ST-SU-PR} & \textbf{ST-SQ-PR} & \\
\midrule
\multicolumn{16}{l}{\textit{\textcolor{gray}{General MLLMs}}} \\
GPT-5.4 & 0.412 & 0.416 & \textbf{0.516} & \textbf{0.516} & 0.416 & 0.384 & 0.436 & 0.416 & \textbf{0.456} & \textbf{0.484} & \textbf{0.424} & 0.384 & 0.400 & 0.416 & \textbf{0.439} \\
GPT-5-chat & 0.424 & 0.400 & 0.472 & 0.316 & 0.276 & 0.352 & 0.380 & 0.368 & 0.388 & 0.388 & 0.388 & 0.276 & 0.280 & 0.292 & 0.356 \\
GPT-4o-2024-11-20 & 0.376 & 0.432 & 0.420 & 0.332 & 0.360 & 0.352 & 0.400 & 0.364 & 0.286 & 0.334 & 0.416 & 0.352 & 0.284 & 0.340 & 0.361 \\
GPT-4o-mini & \textbf{0.428} & 0.416 & 0.428 & 0.328 & 0.400 & 0.356 & 0.352 & 0.336 & 0.248 & 0.304 & \textbf{0.424} & 0.372 & 0.280 & 0.304 & 0.355 \\
Gemini-3.1-Pro-Preview & 0.420 & \textbf{0.448} & 0.496 & 0.396 & \textbf{0.476} & \textbf{0.436} & \textbf{0.460} & \textbf{0.456} & 0.428 & 0.458 & 0.400 & 0.360 & \textbf{0.428} & \textbf{0.432} & 0.436 \\
Gemini-2.5-flash & 0.200 & 0.188 & 0.264 & 0.240 & 0.188 & 0.160 & 0.116 & 0.160 & 0.232 & 0.240 & 0.164 & 0.168 & 0.116 & 0.152 & 0.190 \\
Claude-3.5-haiku & 0.308 & 0.316 & 0.304 & 0.360 & 0.192 & 0.208 & 0.232 & 0.200 & 0.372 & 0.370 & 0.208 & 0.224 & 0.248 & 0.168 & 0.270 \\
Claude-Opus-4.6 & 0.272 & 0.264 & 0.372 & 0.408 & 0.452 & 0.392 & 0.336 & 0.300 & 0.398 & 0.348 & 0.264 & 0.320 & 0.396 & 0.340 & 0.350 \\
Grok-2-vision & 0.188 & 0.216 & 0.288 & 0.368 & 0.232 & 0.240 & 0.240 & 0.280 & 0.252 & 0.300 & 0.220 & 0.196 & 0.172 & 0.148 & 0.240 \\
\cmidrule[0.4pt](lr){1-1} \cmidrule(lr){2-9} \cmidrule(lr){10-11} \cmidrule(lr){12-15}  \cmidrule(lr){16-16}
Deepseek-vl2 & 0.372 & 0.392 & 0.452 & 0.344 & 0.144 & 0.216 & 0.200 & 0.148 & 0.416 & 0.446 & 0.096 & 0.128 & 0.064 & 0.080 & 0.250 \\
GLM-4.5v & 0.268 & 0.136 & 0.248 & 0.312 & 0.152 & 0.084 & 0.084 & 0.132 & 0.312 & 0.340 & 0.132 & 0.172 & 0.180 & 0.112 & 0.190 \\
LLama-3.2-11B & 0.232 & 0.228 & 0.244 & 0.264 & - & - & - & - & 0.292 & 0.286 & - & - & - & - & 0.110 \\
LLama-3.2-90B & 0.368 & 0.364 & 0.356 & 0.308 & 0.236 & 0.268 & 0.300 & 0.268 & 0.408 & 0.430 & 0.308 & 0.352 & 0.268 & 0.212 & 0.318 \\
Qwen2.5-VL-7B & 0.256 & 0.172 & 0.384 & 0.176 & 0.224 & 0.324 & 0.248 & 0.216 & 0.198 & 0.238 & 0.328 & 0.280 & 0.232 & 0.212 & 0.249 \\
Qwen2.5-VL-32B & 0.292 & 0.312 & 0.368 & 0.296 & 0.244 & 0.300 & 0.256 & 0.236 & 0.370 & 0.308 & 0.276 & 0.284 & 0.236 & 0.156 & 0.281 \\
Qwen2.5-VL-72B & 0.216 & 0.300 & 0.296 & 0.392 & 0.316 & 0.240 & 0.216 & 0.204 & 0.402 & 0.370 & 0.172 & 0.180 & 0.204 & 0.188 & 0.264 \\
Qwen3-VL-2B & 0.350 & 0.361 & 0.468 & 0.185 & 0.204 & 0.428 & 0.412 & 0.336 & 0.287 & 0.260 & 0.380 & 0.312 & 0.304 & 0.312 & 0.330 \\
Qwen3-VL-8B & 0.341 & 0.363 & 0.472 & 0.211 & 0.388 & 0.405 & 0.398 & 0.371 & 0.233 & 0.275 & 0.381 & 0.401 & 0.401 & 0.384 & 0.359 \\
Qwen3-VL-32B & 0.372 & 0.416 & 0.476 & 0.316 & 0.408 & 0.428 & 0.388 & 0.392 & 0.416 & 0.364 & 0.388 & \textbf{0.456} & 0.336 & 0.372 & 0.395 \\
\midrule
\multicolumn{16}{l}{\textit{\textcolor{gray}{Remote Sensing MLLMs}}} \\
GeoChat & 0.280 & 0.332 & 0.360 & 0.308 & - & - & - & - & 0.352 & 0.356 & - & - & - & - & 0.331 \\
VHM & 0.302 & 0.297 & 0.308 & 0.210 & - & - & - & - & 0.324 & 0.332 & - & - & - & - & 0.296 \\
ScoreRS w/ SFT & 0.403 & 0.367 & 0.421 & 0.345 & 0.294 & 0.310 & 0.288 & 0.284 & 0.382 & 0.419 & 0.341 & 0.320 & 0.313 & 0.295 & 0.347 \\
ScoreRS w/ RL & 0.313 & 0.338 & 0.382 & 0.295 & 0.399 & 0.335 & 0.367 & \textbf{0.392} & 0.409 & 0.371 & 0.338 & 0.313 & 0.382 & 0.342 & 0.355 \\
\midrule
\textbf{Base }           & 0.300 & 0.300 & 0.300 & 0.316 & 0.188 & 0.268 & 0.196 & 0.224 & 0.306 & 0.300 & 0.348 & 0.280 & 0.268 & 0.236 & 0.274 \\
\textbf{Best eval (2k step) } & 0.296 & 0.268 & 0.272 & 0.300 & 0.196 & 0.328 & 0.272 & 0.312 & 0.276 & 0.294 & 0.336 & 0.320 & 0.296 & 0.260 & 0.288 \\
\textbf{MLRS (5k step) }      & 0.248 & 0.180 & 0.224 & 0.240 & 0.176 & 0.232 & 0.236 & 0.244 & 0.250 & 0.250 & 0.236 & 0.264 & 0.252 & 0.188 & 0.230 \\
\bottomrule
\end{tabular}
}
\label{tab:model_performance}
\end{table*}

\begin{table*}[ht!]
  \centering
    \caption{UrBench~\cite{zhou2024urbench} VQA. Left-to-right order: CR=City Retrieval, IR=Image Retrieval, CL-Camera Localization, OR=Orientation, SR=Scene Recognition, RU=Road Understanding, CO=Counting, SC=Scene Comparison, RBR=Role-based Reasoning, TSR=Traffic Sign Reasoning, VPR=Visual Prompt Reasoning, OM=Object Matching, OG=Object Grounding, OAR=Object Attribute Recognition. Column-wise maxima are shown in bold.}
\label{tab:pt_ur}
  \resizebox{1\textwidth}{!}{
    \begin{tabular}{l|cccc|cccc|ccc|ccc|c}
    \toprule
    \multirow{1}[4]{*}{\textbf{Model}} 
    & \multicolumn{4}{c|}{\shortstack{\textbf{Geo-} \\ \textbf{Localization}}}
    & \multicolumn{4}{c|}{\shortstack{\textbf{Scene} \\ \textbf{Understanding}}}
    & \multicolumn{3}{c|}{\shortstack{\textbf{Scene} \\ \textbf{Reasoning}}}
    & \multicolumn{3}{c|}{\shortstack{\textbf{Object} \\ \textbf{Understanding}}} 
    & \multirow{1}[4]{*}{\textbf{Overall}} \\
\cmidrule{2-15}          & \textbf{CR} & \textbf{IR} & \textbf{CL} & \textbf{OR} & \textbf{SR} & \textbf{RU} & \textbf{CO} & \textbf{SC} & \textbf{RBR} & \textbf{TSR} & \textbf{VPR} & \textbf{OM} & \textbf{OG} & \textbf{OAR} &  \\
    \midrule
    Human & 30.0  & 92.6  & 82.9  & 85.7  & 59.2  & 87.2  & 94.1  & 85.1  & 87.4  & 85.7  & 88.2  & 95.2  & 95.5  & 61.6  & 69.9  \\
    Random & 24.8  & 23.9  & 25.1  & 23.2  & 17.7  & 25.7  & 21.4  & 25.3  & 23.9  & 24.2  & 30.6  & 21.8  & 22.1  & 21.5  & 23.5  \\
    \midrule
    GPT-4o & \textbf{79.2 } & \textbf{85.9 } & {35.3}  & {30.7}  & \textbf{65.0 } & {66.3}  & 40.1  & {79.0}  & \textbf{79.6} & \textbf{68.2}  & \textbf{77.9} & 28.0  & 46.5  & \textbf{50.1} & \textbf{61.2} \\
    Gemini-1.5-Flash & 69.7  & 25.9  & 25.9  & 24.0  & 57.9  & \textbf{71.0}  & 29.1  & 67.7  & 77.8  & \textbf{75.8} & 69.8  & 22.0  & 39.1  & 40.9  & 50.9  \\
    Claude-3.5-Sonnet & {72.3}  & 55.8  & 30.8  & \textbf{33.3}  & 52.4  & 59.0 & \textbf{48.0 } & \textbf{81.0 } & 73.7  & 37.7  & 66.7  & 22.0  & \textbf{61.5 } & 45.4  & {55.0}  \\
    \midrule
    TinyLLaVA & 51.9  & 23.2  & 24.7  & 27.9  & 8.6   & 9.3  & 9.5  & 27.6  & 40.3  & 32.7  & 48.6  & 22.9  & 41.1  & 18.8  & 29.9  \\
    InternVL2-2B & 50.3  & 23.8  & 31.9  & 29.0  & 47.9  & 47.9  & 28.6  & 30.1  & 64.8  & 45.9  & 54.5  & 25.5  & 30.3  & 40.5  & 41.2  \\
    InternVL2-4B & 55.0  & 24.2  & 27.4  & 23.1  & 52.3  & 53.1  & 22.1  & 39.4  & 73.0  & 56.6  & 62.6  & 30.6  & 30.2  & 40.2  & 43.5  \\
    XComposer2-4KHD & 61.9  & 26.0  & 27.5  & 25.9  & 55.6  & 65.8  & 35.8  & 30.3  & 75.5  & 62.6  & 60.8  & 16.2  & 42.6  & 46.9  & 47.8  \\
    LLaVA-NeXT-7B-Mistral & 51.6  & 25.9  & 24.2  & 24.0  & 55.6  & 47.3  & 34.1  & 28.4  & 59.2  & 42.7  & 45.5  & \textbf{43.9} & 33.1  & 30.0  & 39.2  \\
    LLaVA-NeXT-7B-Vicuna & 51.2  & 24.6  & 27.2  & 23.3  & 56.1  & 20.2  & 34.3  & 25.9  & 49.6  & 51.5  & 54.5  & 31.8  & 27.2  & 31.9  & 37.1  \\
    InstructBLIP-Vicuna-7B & 40.4  & 25.7  & 25.4  & 27.1  & 33.0  & 14.8  & 20.4  & 28.0  & 30.7  & 25.7  & 29.3  & 22.9  & 22.7  & 17.3  & 27.6  \\
    LLaVA-NeXT-Interleave-7B & 57.9  & 41.6  & 27.6  & 25.5  & 52.6  & 50.7  & 37.3  & 48.4  & 65.8  & 55.9  & 63.1  & 37.3  & 37.6  & 41.5  & 40.4  \\
    Mantis-LLaMA3-SigLIP & 67.0  & 32.4  & 27.0  & 27.2  & {59.2}  & 44.5  & 27.4  & 52.4  & 67.6  & 41.6  & 57.7  & 25.2  & 34.2  & 38.6  & 45.3  \\
    Mantis-Idefics2 & 69.0  & 29.9  & 27.0  & 25.7  & 50.3  & 49.3  & 22.9  & 56.0  & 68.9  & 50.6  & 56.3  & 29.6  & 37.8  & 35.6  & 40.7  \\
    LLaVA-NeXT-8B & 54.4  & 27.0  & 27.8  & 26.0  & 55.7  & 44.5  & 34.1  & 24.0  & 55.2  & 52.8  & 58.6  & {39.8}  & 41.7  & 28.4  & 38.8  \\
    InternVL2-8B & 50.8  & 26.6  & 31.8  & 25.2  & 53.0  & 52.6  & {43.0}  & 51.4  & 74.9  & 54.8  & 62.6  & 30.3  & 32.0  & 41.3  & 48.8  \\
    Idefics-2-8B & 65.5  & 23.8  & 26.0  & 24.1  & 52.3  & 47.9  & 25.9  & 27.8  & 64.7  & 60.4  & 42.8  & 24.8  & 21.1  & 27.4  & 42.7  \\
    LLaVA-NeXT-13B & 52.0  & 24.5  & 27.7  & 26.7  & 53.9  & 50.7  & 33.8  & 25.1  & 54.0  & 52.1  & 52.3  & 31.8  & 34.8  & 26.3  & 46.5  \\
    VILA-1.5-13B & 62.7  & 33.7  & 28.6  & 24.1  & 47.7  & 43.9  & 23.9  & 48.6  & 66.3  & 43.8  & 46.4  & 25.8  & 32.3  & 38.2  & 45.8  \\
    InternVL2-26b & 61.3  & 23.0  & 32.3  & 24.7  & \textbf{65.0 } & 41.1  & 30.1  & 52.6  & 77.9  & 63.8  & 71.2  & 26.1  & 37.3  & 48.4  & 46.0 \\
    LLaVA-NeXT-34B & 58.4  & 26.0  & 28.5  & 27.8  & 58.3  & 49.0  & 21.6  & 53.9  & 65.6  & 59.3  & 56.8  & 28.7  & 40.5  & 24.1  & 43.7  \\
    VILA-1.5-40B & 70.1  & {62.5}  & \textbf{36.8 } & 27.9  & 53.6  & 51.7  & 32.1  & 66.7  & 76.4  & 55.5  & 61.3  & 34.1  & {48.3}  & 39.5  & 53.1  \\
    \midrule
    \textbf{Base} & 54.2 & 34.7 & 29.9 & 23.6 & 95.8 & 53.1 & 31.1 & 60.0 & 75.3 & 64.2 & 72.5 & 29.2 & 31.4 & 42.6 & 49.8 \\
    \textbf{MLRS} & 54.8 & 42.6 & 26.9 & 22.9 & 95.4 & 52.1 & 34.8 & 59.4 & 78.6 & 66.9 & 76.1 & 32.9 & 35.2 & 47.4 & 51.9 \\
    \bottomrule
    \end{tabular}%
    }
\end{table*}%

\begin{table*}[ht!]
    \centering
    \caption{LaSeRS~\cite{xin2025segearthr2} segmentation. Metric is IoU/cIoU Column-wise maxima are shown in bold.}
\label{tab:pt_lasers}
    \resizebox{\textwidth}{!}{
\begin{tabular}{l|ccc|cc|cc|cc|c}
    \toprule
    \multirow{2}{*}{Model} & \multicolumn{3}{c|}{{Hierarchical Segmentation Granularity}} & \multicolumn{2}{c|}{{Target Multiplicity}} & \multicolumn{2}{c|}{{Reasoning Requirements}} & \multicolumn{2}{c|}{{Linguistic Variability}} & \multirow{2}{*}{Avg.} \\
     & {Semantic} & {Instance} & {Part} & {Single} & {Multiple} & {Explicit} & {Implicit} & {Short} & {Long} & \\
    \midrule
    LISA-7B & 26.4/23.2 & 20.5/25.0 & 16.1/11.6 & 37.3/32.2 & 18.2/22.4 & 27.1/24.3 & 21.5/25.6 & 34.1/27.8 & 38.4/33.9 & 26.6/25.1 \\
    LISA-13B & 27.0/24.5 & 22.3/25.6 & 17.7/13.1 & 38.4/34.2 & 19.9/23.5 & 27.1/25.5 & 22.6/25.8 & 35.2/28.0 & 38.4/34.3 & 27.6/26.1 \\
    PixelLM-7B & 32.0/32.8 & 26.6/30.0 & 13.2/16.5 & 44.3/40.4 & 20.2/23.5 & 25.0/23.1 & 23.9/21.9 & 41.6/38.9 & 37.1/34.5 & 29.3/29.1 \\
    PixelLM-13B & 31.6/34.0 & 27.5/30.2 & 15.8/17.6 & 42.2/40.5 & 20.9/22.4 & 26.3/24.4 & 25.9/22.1 & 42.0/39.1 & 37.1/34.5 & 29.9/29.4 \\		
    GLaMM-ft-7B & 44.8/47.9 & 41.2/48.3 & 32.6/42.7 & 47.3/{50.3} & 32.2/41.0 & 59.1/60.3 & {42.6}/44.8 & 50.4/54.8 & 42.6/44.8 & 43.6/48.3 \\
    $M^2$A-7B & 30.1/33.0 & 23.0/24.8 & 18.6/17.2 & 45.4/37.6 & 20.9/24.8 & 35.8/30.4 & 23.3/26.7 & 35.8/32.8 & 41.5/36.7 & 30.5/29.3 \\
    GeoPixel-8B & {51.4}/{57.2} & {44.1}/{49.3} & {43.9}/{52.4} & {55.0}/45.8 & {49.2}/{49.7} & {66.5}/{61.3} & 41.1/{58.3} & {51.1}/{59.3} & {51.4}/{63.2} & {50.4}/{55.2} \\
    {SegEarth-R2-3B} & {60.2/\textbf{71.8}} & {{65.4}/\textbf{70.3}} & {\textbf{64.8}/\textbf{68.3}} & {{55.1}/\textbf{69.2}} & {{38.3}/{56.2}} & {\textbf{78.4}/\textbf{80.4}} & {{42.8}/{59.7}} & {{60.2}/{69.9}} & {{50.1}/{65.7}} & {{57.2}/\textbf{67.9}} \\
    \midrule
    \textbf{Base} & 40.2/39.4 & 41.3/32.6 & 15.4/19.7 & 37.9/35.1 & 1.3/1.3 & 33.2/27.4 & 30.7/23.7 & 36.4/45.8 & 42.9/35.6 & 31.1/29.0 \\
    \textbf{MLRS} & \textbf{71.1}/70.7 & \textbf{72.2}/70.0 & 51.3/63.1 & \textbf{67.3}/59.2 & \textbf{59.5}/\textbf{58.7} & 67.4/65.6 & \textbf{68.1}/\textbf{67.0} & \textbf{70.1}/\textbf{74.5} & \textbf{75.2}/\textbf{77.9} & \textbf{66.9}/67.4 \\
    \bottomrule
\end{tabular}
    }
\end{table*}

\begin{table}[!ht]
  \caption{GeoSeg-Bench~\cite{ni2025unigeoseg} segmentation. The best results are shown in bold.}
\label{tab:pt_gsb}
  \centering
  \resizebox{.6\textwidth}{!}{
  \begin{tabular}{@{}lcccccc}

    \toprule
    \multirow{2}{*}{Method} & \multicolumn{2}{c}{Interactive} & \multicolumn{2}{c}{Referring} & \multicolumn{2}{c}{Reasoning} \\
    \cmidrule(lr){2-3} \cmidrule(lr){4-5} \cmidrule(lr){6-7}
    & cIoU & gIoU & cIoU & gIoU & cIoU & gIoU \\
    \midrule
    LISA & 2.52 & 3.12 & 3.53 & 4.56 & 7.09 & 5.77 \\
    PixelLM & 0.08 & 0.11 & 6.04 & 5.80 & 6.36 & 6.30 \\
    PSALM & 6.35 & 10.83  & 31.77 & 18.91 & 11.88 & 9.27 \\
    HIPIE & 6.01 & 12.06 & 29.25 & 39.14 & 7.90 & 11.76 \\
    SegLLM & 8.97 & 17.72 & 15.90 & 35.27 & 11.06 & 16.41 \\
    Geopixel & 17.21 & 18.71  & 37.34 & 40.14 & 27.36 & 26.71 \\
    Geopix & 15.28 & 17.63 & 28.81 & 28.52 & 20.80 & 18.25 \\
    RemoteSAM & 4.85 & 6.49 & 12.06 & 29.31 & 8.09 & 9.01 \\
    Earthmind & 16.38 & 16.57 & 44.53 & 46.48 & 31.01 & 28.80 \\
    LISAT & 4.27 & 5.52 & 37.87 & 40.14 & 22.06 & 20.09 \\
    Segearth-R1 & 4.88 & 5.28 & 14.65 & 9.39 & 12.72 & 11.97 \\
    \midrule
    PSALM-ft & 70.78 & 74.10 & 68.70 & 71.15 & 47.53 & 49.59 \\
    Geopixel-ft  & 42.62 & 46.48 & 42.70 & 45.50 & 30.25 & 28.51 \\
    Earthmnind-ft  & 67.74 & 70.89 & 48.09 & 49.24 & 36.84 & 25.71 \\
    LISAT-ft  & 68.43 & 73.00 & 59.82 & 62.46 & 41.53 & 31.25 \\
    Segearth-R1-ft  & {72.09} & {75.00} & {70.76} & {72.98} & {53.31} & {51.56} \\
    UniGeoSeg-ft  & \textbf{74.44} & \textbf{75.56} & \textbf{72.93} & \textbf{74.58} & \textbf{58.35} & \textbf{53.12} \\
    \midrule
    \textbf{Base} & 43.6 & 43.5 & 19.7 & 18.2 & 21.2 & 16.9 \\
    \textbf{MLRS} & 42.3 & 46.5 & 53.0 & 57.0 & 35.0 & 37.7 \\
    \bottomrule
  \end{tabular}
  }
\end{table}

\section{More example outputs}

\begin{figure}
    \centering
    \includegraphics[width=1\linewidth]{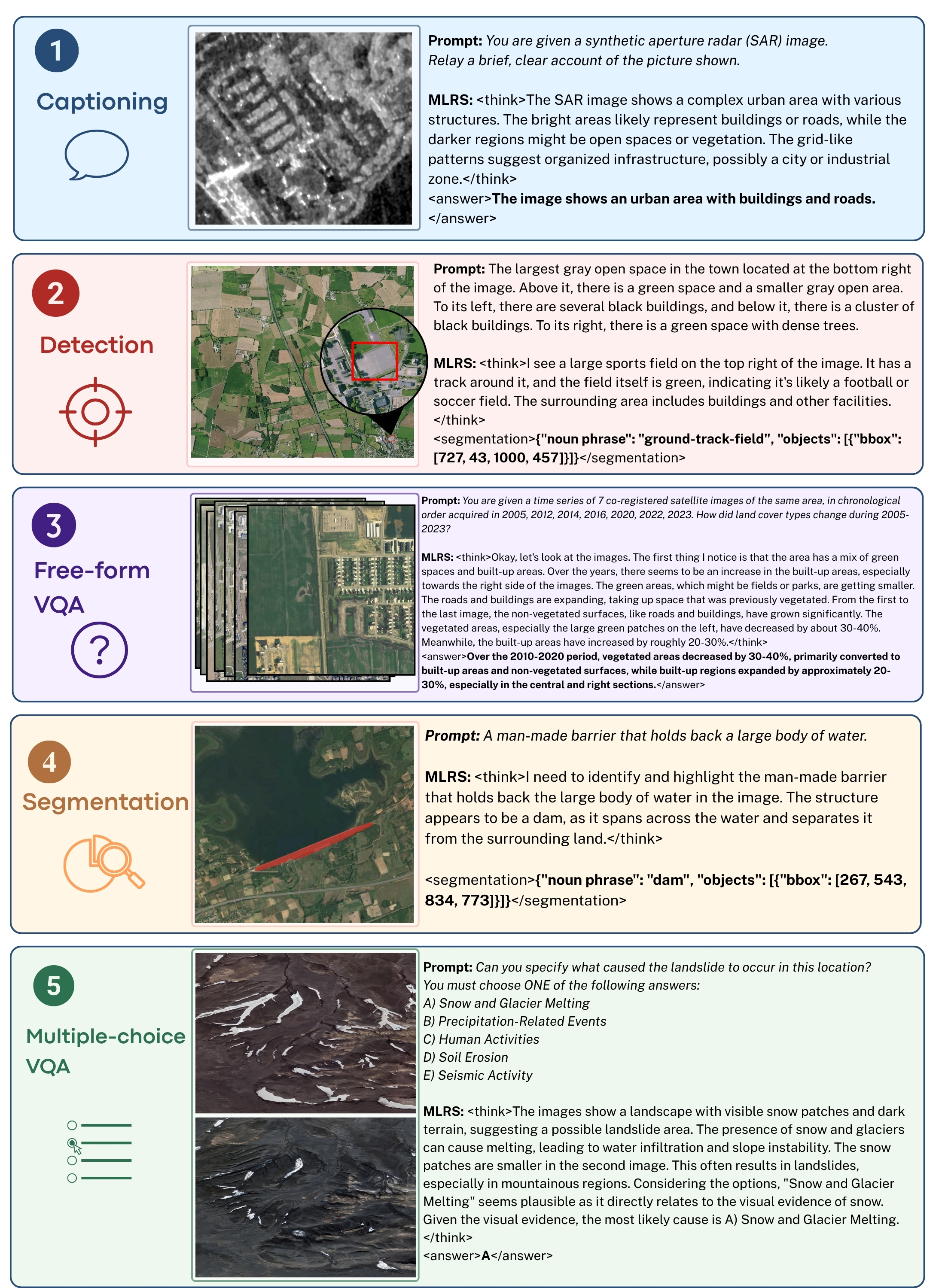}
    \caption{Example outputs of MLRS for five general tasks: captioning, detection, free-form VQA, segmentation, multiple-choice VQA. For detection, the bounding box is visualized on the high-resolution input image and zoomed in. For segmentation, SAM3 is prompted with the VLM's output, and the resulting mask is visualized on the input image.}
    \label{fig:example_outputs_app1}
\end{figure}

\begin{figure}
    \centering
    \includegraphics[width=1\linewidth]{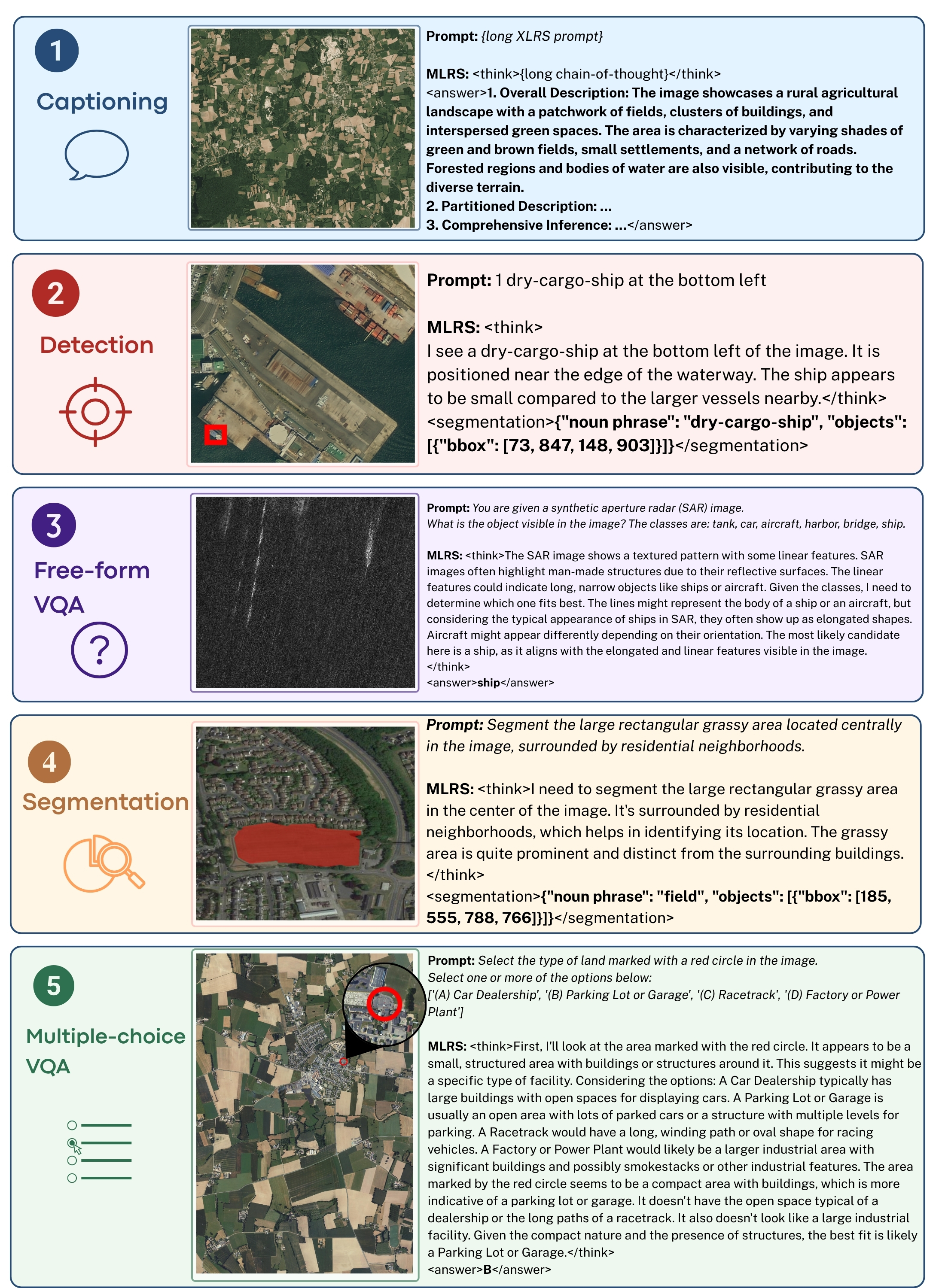}
    \caption{Example outputs of MLRS for five general tasks: captioning, detection, free-form VQA, segmentation, multiple-choice VQA. The captioning prompt and model output are shortened. For detection, the bounding box is visualized on the input image. For segmentation, SAM3 is prompted with the VLM's output, and the resulting mask is visualized on the input image. The MC VQA example prompts visually: the target is zoomed in the high-resolution input image.}
    \label{fig:example_outputs_app2}
\end{figure}

Additional examples of MLRS outputs are provided in Figures~\ref{fig:example_outputs_app1} and~\ref{fig:example_outputs_app2}.

\end{document}